\documentclass[10pt, a4paper]{article}
\usepackage{hybridreport}
 
\usepackage{cleveref}
\usepackage{wrapfig}
\usepackage{threeparttable}
\usepackage{multirow}
\usepackage{tabularx}
\usepackage{makecell}
\usepackage{wasysym} \usepackage{xspace}
\usepackage{tikz}

\usepackage[square,authoryear]{natbib}

\def\DBperiod{six months\xspace}
\def\DBNtask{1300\xspace}
\def\DBhours{54,560\xspace}
\def\DBhoursPT{8,434\xspace} \def\DBep{5.56M\xspace} \def\AvgSubTask{6.02\xspace} \def\AvgEpTime{35.4\xspace} \def\DBdemo{25.5M\xspace}
\def\Nuser{363\xspace}
\def\Nuserm{283\xspace} \def\Nuserf{80\xspace} \def\Ndesk{44\xspace}

\def\tbf{\textbf}

\makeatletter
\DeclareRobustCommand\onedot{\futurelet\@let@token\@onedot}
\def\@onedot{\ifx\@let@token.\else.\null\fi\xspace}
\def\eg{\emph{e.g}\onedot} 
\def\ie{\emph{i.e}\onedot}

\makeatother

\makeatletter
\newcommand{\customparagraph}[1]{\par\noindent\textbf{#1:}}
\makeatother

\newcommand{\hbFull}{\mbox{\begin{tikzpicture}[baseline=-0.5ex]
        \draw[thick] (0,0) circle (0.7ex);
        \fill[black] (0,0) -- (90:0.7ex) arc (90:-270:0.7ex) -- cycle;
    \end{tikzpicture}}}
\newcommand{\hbHalf}{\mbox{\begin{tikzpicture}[baseline=-0.5ex]
        \draw[thick] (0,0) circle (0.7ex);
        \fill[black] (0,0) -- (90:0.7ex) arc (90:-90:0.7ex) -- cycle;
    \end{tikzpicture}}}
\newcommand{\hbEmpty}{\mbox{\begin{tikzpicture}[baseline=-0.5ex]
        \draw[thick] (0,0) circle (0.7ex);
    \end{tikzpicture}}}

\newcommand{\hdashline}{\noalign{\vskip1.5pt}
  \noalign{\hbox{\tikz[baseline]{\draw[densely dashed, line width=0.35pt] (0,0) -- (\linewidth,0);}}}
  \noalign{\vskip1.5pt}}

\ifdefined\paperreview
   \else
  
\fi

\newcommand{\todo}[1]{}
\newcommand{\konote}[1]{}

\DeclareTextFontCommand{\textbf}{\bfseries}
\providecommand{\eg}{\emph{e.g.}\xspace}

\providecommand{\ie}{\emph{i.e.}\xspace}

\providecommand{\tbf}[1]{\textbf{#1}}
\providecommand{\customparagraph}[1]{\par\noindent\textbf{#1:}}

\IfFontExistsTF{Hiragino Mincho ProN}{\newfontfamily{\jpfont}{Hiragino Mincho ProN}}{\newcommand{\jpfont}{}}

\reporttype{Technical Report}
\title{YUBI: Yielding Universal Bidigital Interface for Bimanual Dexterous Manipulation at Scale}
\shorttitle{YUBI: Yielding Universal Bidigital Interface}

\author{
  \mbox{Takehiko Ohkawa\textsuperscript{1*}}\quad 
  \mbox{Jumpei Arima\textsuperscript{2*}}\quad
  \mbox{Yuki Noguchi\textsuperscript{1}}\quad
  \mbox{Masatoshi Tateno\textsuperscript{1,4}}\quad
  \mbox{Makoto Sugiura\textsuperscript{1}}\quad
  \mbox{Takuya Okubo\textsuperscript{1}}\quad
  \mbox{Kengo Ikeuchi\textsuperscript{1,4}}\quad
  \mbox{Yuma Shin\textsuperscript{1,5}}\quad
  \mbox{Hiroki Nishizawa\textsuperscript{1,6}}\quad
  \mbox{Naoaki Kanazawa\textsuperscript{1}}\quad
  \mbox{Yuki Wakayama\textsuperscript{2}}\quad
  \mbox{Daiki Fukunaga\textsuperscript{2}}\quad
  \mbox{Koshi Makihara\textsuperscript{3}}\quad
  \mbox{Tomohiro Motoda\textsuperscript{3}}\quad
  \mbox{Floris Erich\textsuperscript{3}}\quad
  \mbox{Yukiyasu Domae\textsuperscript{3}}\quad
  \mbox{Tatsuya Matsushima\textsuperscript{1,4}}\quad
  \mbox{Yoshihiro Okumatsu\textsuperscript{2}}\quad
  \mbox{Kei Ota\textsuperscript{1}}
}

\institution{
  \textsuperscript{1}AI Robot Association (AIRoA)\quad
  \textsuperscript{2}Toyota Motor Corporation\\
  \textsuperscript{3}National Institute of Advanced Industrial Science and Technology (AIST)\\
  \textsuperscript{4}The University of Tokyo\quad
  \textsuperscript{5}Institute of Science Tokyo\quad
  \textsuperscript{6}Waseda University
}

\date{November 2026}
\correspondence{\href{mailto:ohkawa.takehiko@airoa.org}{\texttt{ohkawa.takehiko@airoa.org}}\\
  Project page: \href{https://yubi.airoa.io/}{\texttt{https://yubi.airoa.io/}}\\
  \textsuperscript{*}Project Co-Lead}

\abstracttext{
  We introduce \emph{Yielding Universal Bidigital Interface (YUBI)}, a finger-aligned gripper designed to enable intuitive, ergonomic, and scalable data collection for bimanual dexterous manipulation. While handheld data collection systems such as Universal Manipulation Interface (UMI) enable affordable data collection, their bulky pistol-grip designs can pose ergonomic and usability challenges for fine-grained, dexterous manipulation tasks. To address this, YUBI presents a distinct design principle: \emph{yielding, finger-driven actuation} that directly maps human finger movements to gripper jaw motion. Using the YUBI devices, we set up a data collection system with integrated VR-based 6\,DoF tracking of the gripper, ensuring high-fidelity trajectory data acquisition. We curate a UMI-based dataset of unprecedented scale: \DBhours hours across \DBep episodes and \DBNtask tasks. Experiments show that YUBI offers advantages over the UMI gripper in versatility for complex bimanual tasks, dexterity, and operational efficiency. A single policy trained on the YUBI dataset transfers across multiple bimanual robots (UR, Franka, and ELEY) simply by mounting the gripper on each platform, confirming that the collected data are directly executable as policy supervision. We release the gripper hardware, data-collection software, and dataset as one integrated stack, offering the open community a reproducible path to large-scale data acquisition for advancing robotic foundation models.
}

\begin{document}
\maketitle

\begin{figure}[t]
    \centering
    \includegraphics[width=\linewidth]{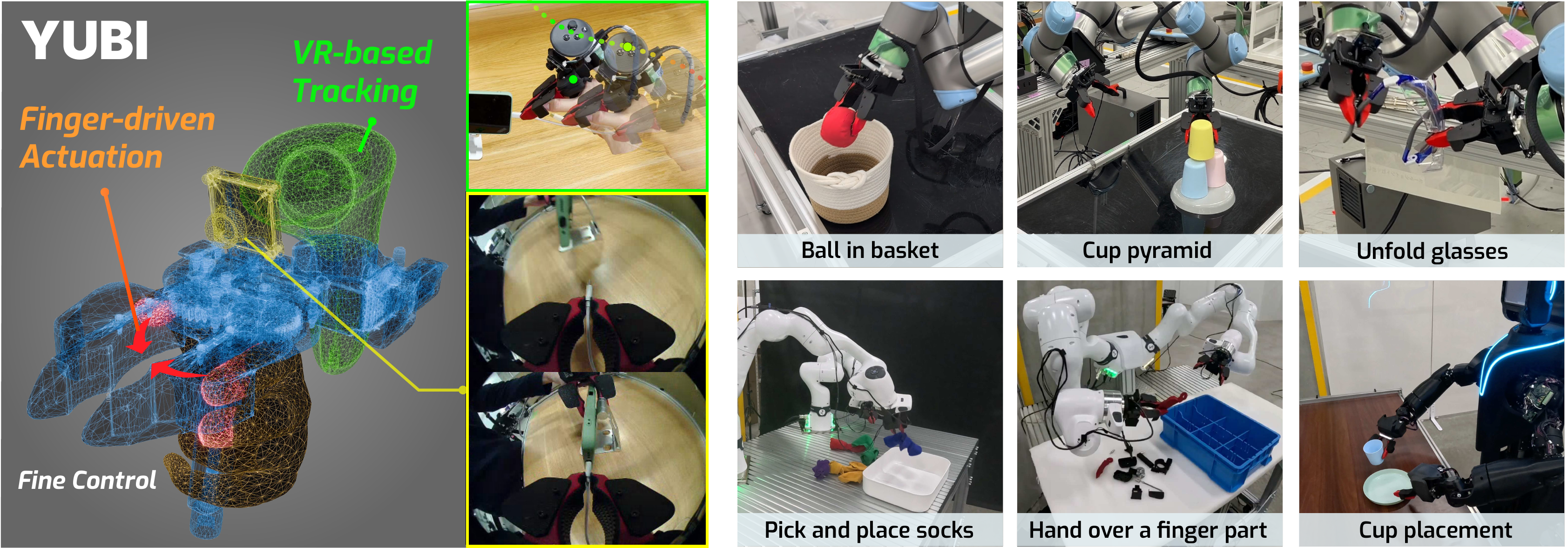}
    \caption{\textbf{Yielding Universal Bidigital Interface (YUBI):}
    Our lightweight, finger-aligned gripper offers intuitive control by mirroring human digital kinematics for dexterous manipulation. 
    YUBI facilitates the curation of a large-scale, high-quality bimanual dataset for \textit{cross-embodiment VLA deployment} on multiple robot platforms (UR, Franka, ELEY) without robot-specific tuning.   
    }
    \label{fig:teaser}
\end{figure}

\section{Introduction}
Scaling data collection is fundamental to the progress of robotic foundation models. 
As demonstrated by the success of large multimodal models, the performance and generalization of Vision-Language-Action (VLA) policies are bounded by the volume and diversity of their training data~\cite{black2024pi0,black2025pi05,o2024open,generalistai2025gen0}.
Yet, a critical bottleneck remains: while frontier industry labs leverage massive, proprietary datasets to produce the most capable VLA models~\cite{generalistai2026,sundayrobotics2026}, the open research community is left without accessible demonstrations at a comparable scale.
We set out to close this gap.

Fundamentally, current data collection systems struggle to supply large-scale dexterous manipulation data at high throughput.
Traditional leader-follower teleoperation, while providing high-fidelity robot-space demonstrations, is cost-prohibitive and requires operator expertise, resulting in low data throughput~\cite{zhao2023learning,wu2024gello,buamanee2024bi}.
Conversely, imitation learning from human demonstrations offers a more scalable alternative.
While raw human demonstration data (\eg, via web videos or wearable sensors) are available at scale~\cite{grauman2022ego4d,punamiya2026egoverse,ohkawa2023assemblyhands}, they introduce the challenge of a fundamental embodiment gap between human hands and robotic hardware. 
Recently, handheld interfaces like Universal Manipulation Interface (UMI)~\cite{chi2024universal,liu2025fastumi,zhaxizhuoma2025fastumi} have emerged as a pivotal medium between humans and robots. 
These UMI systems bridge the embodiment gap by having operators guide a gripper mechanically identical to the robot's end-effector.
This eliminates the need for physical robots during data collection
and provides wrist camera views and trajectory data for policy learning.
Such UMI-based systems suggest a cost-effective and embodiment-agnostic pathway for large-scale data collection.

\begin{table}[t]
\centering
\renewcommand{\arraystretch}{1.15}
\caption{Comparative analysis of UMI systems.}
\label{table:umi_comparison}
\begin{threeparttable}
    \footnotesize
    \begin{tabularx}{\linewidth}{@{}l XXX@{}}
    \toprule
    \footnotesize\makecell[lt]{\textbf{Method}}
  & \footnotesize\makecell[lt]{\textbf{UMI~\cite{chi2024universal}} \\
                               \textbf{FastUMI~\cite{zhaxizhuoma2025fastumi}}}
  & \footnotesize\makecell[lt]{\textbf{ActiveUMI~\cite{zeng2025activeumi}} \\
                               \textbf{exUMI~\cite{xu2025exumi}}}
  & \footnotesize\makecell[lt]{\textbf{YUBI} \\ \textbf{(Ours)}} \\
    \midrule
    Gripper          & Pistol-grip            & Pistol-grip               & \emph{Finger-driven}      \\
    Weight           & 780\,g                 & ($\ge$\,905\,g)           & 319\,g                    \\
    Ergonomics       & \hbHalf~(bulky)        & \hbEmpty~(heavy HMD)      & \hbFull~(haptic\,$\uparrow$) \\
    Dexterity        & \hbHalf                & \hbHalf                   & \hbFull                   \\ \hdashline
    Tracking         & SLAM                   & VR                        & VR                        \\
    Tracking prec.   & \hbHalf                & \hbFull                   & \hbFull                   \\ \hdashline
    Data size (hrs)  & 12 / (60)              & -- / 5                    & \DBhours                  \\
    \#Tasks          & 4 / 22                 & 6 / 9                     & \DBNtask                  \\
    \bottomrule
    \end{tabularx}
    \begin{tablenotes}
        \footnotesize
        \item \hbFull: Superior, \hbHalf: Moderate, \hbEmpty: Limited. Parenthesized values are estimates.
    \end{tablenotes}
\end{threeparttable}
\end{table}
 
Despite the success of initial handheld designs, scaling to increasingly complex tasks reveals ergonomic and functional bottlenecks, as shown in \Cref{table:umi_comparison}.
Most existing UMI frameworks use a pistol-grip design, which introduces a mechanical offset between the operator's fingers and the gripper's pinch-point. 
This offset, combined with the heavy and bulky shell, often obstructs natural hand movements, limiting the manipulability and dexterity of the gripper.
Furthermore, the gripper's tracking fidelity remains a hurdle. 
Unlike SLAM-based tracking in \cite{chi2024universal,liu2025fastumi,lin2024data,liu2025vitamin}, recent systems~\cite{zeng2025activeumi,xu2025exumi} have incorporated Virtual Reality (VR) systems for reliable gripper tracking by mounting a VR headset on the operator.
However, wearing the headset introduces substantial physical strain on the neck, which limits session length and operational efficiency (up to 30\,min of continuous use~\cite{meta2026quest3s_safety}).

To address this, we reinvent the handheld UMI as a finger-aligned interface, called \tbf{YUBI (Yielding Universal Bidigital Interface)}\footnote{\emph{YUBI} is a Japanese reading of the kanji character {\jpfont 指}, meaning ``finger''.}, enabling intuitive and sustained data collection for bimanual dexterous manipulation.
As shown in \Cref{fig:teaser}, YUBI introduces \emph{yielding, finger-driven actuation}, where the gripper aperture follows the operator's natural pinch motion, eliminating the mechanical offset of pistol-grip UMI designs and offering intuitive haptic feedback to the fingers.
Additionally, the sharpened fingertip geometry enables precise, fine-grained control with a parallel-jaw gripper (\eg, box assembly and cable insertion).
Each YUBI gripper can be fabricated for under \$200~USD with 3D printed parts, 
making the design accessible to the research community.

Combined with the gripper redesign, the YUBI system integrates high-frequency VR sensors directly into each gripper and supports both stationary and portable data collection. 
Quest provides a pose for nearly every input frame and has fewer missing poses than the UMI-style monocular ORB-SLAM3 baseline (\Cref{supp:tracking_accuracy}).
To sustain throughput, we attach the VR headset to a stationary camera rig rather than to the operator's head, eliminating physical load.
Conversely, a portable setup allows operation beyond tabletop scenarios,  such as household tasks, with whole-body motion and locomotion.

Leveraging this framework, we provide the largest UMI-based dataset for bimanual dexterous manipulation, comprising \DBhours~hours of data across \DBep~episodes and \DBNtask~tasks. 
Our user study finds that YUBI is adaptable to tasks ranging from daily to industrial scenarios, with improved operational efficiency and task success rates over the original UMI gripper.
We further demonstrate the effectiveness of the YUBI dataset by pre-training and post-training study of a single policy network, and deploying it on three bimanual robot platforms, namely UR~\cite{ur2026}, Franka~\cite{franka2026}, 
and Toyota's semi-humanoid ELEY~\cite{eley2026}, each fitted with the YUBI gripper as a common end-effector.
Leveraging an \DBhoursPT{}-hour subset of the YUBI dataset as the pretraining set, the downstream performance significantly improves from 10\% to 65\% on Sort YUBI parts task.
To address data scarcity in the open research community, we fully release the YUBI ecosystem, encompassing gripper hardware, data collection software, and our large-scale dataset, empowering the community to reproduce and scale dexterous bimanual manipulation research.

Our contributions are summarized as follows:
\begin{itemize}
    \setlength{\parskip}{0pt}
    \setlength{\itemsep}{3pt}
    \vspace{-2mm}
    \item We propose YUBI, a novel finger-aligned gripper with yielding jaw actuation designed to collect interaction data for bimanual dexterous tasks.
    \item We present a VR-based operation setup for both stationary and portable data collection, enabling high-fidelity and sustained data collection.
    \item We offer a large-scale dataset for diverse bimanual dexterous manipulation tasks, comprising \DBhours hours of interaction data across \DBep episodes and \DBNtask tasks.    
    \item We demonstrate the system's efficacy through a comparative user study with the original UMI device, suggesting improved dexterity and operational efficiency.    
    \item We demonstrate that YUBI data provide effective supervision through policy deployment on three bimanual robot platforms and an \DBhoursPT{}-hour pre-training study.    
\end{itemize}

\begin{figure}[t]
    \centering
    \includegraphics[width=\linewidth]{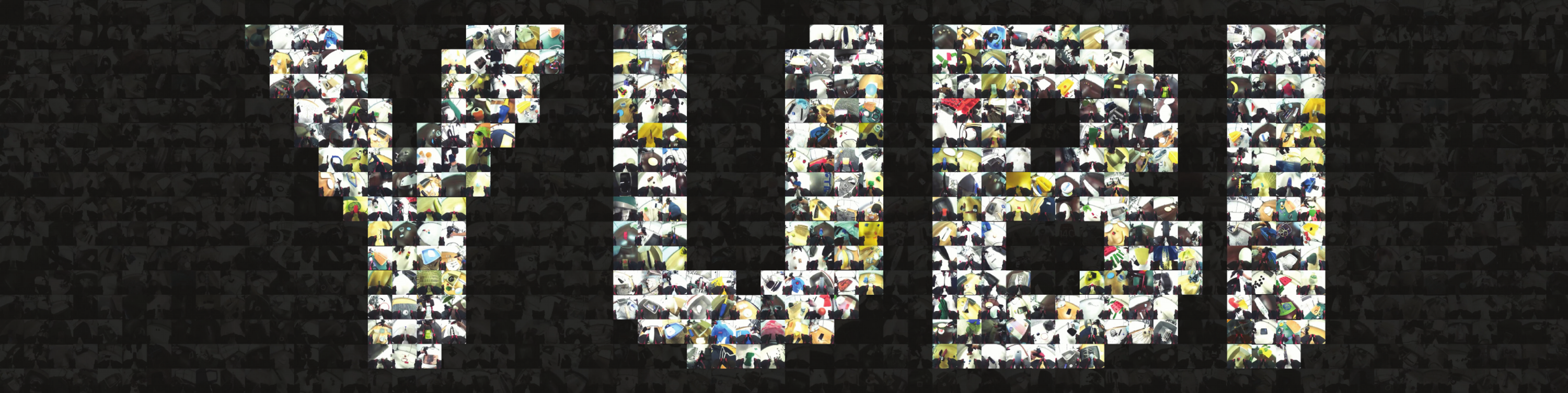}
    \caption{\textbf{YUBI dataset at a glance.} The YUBI dataset spans \DBhours{} hours, \DBdemo{} demos, and \DBNtask{} tasks, covering diverse domains such as kitchen, clothing, retail, and industry, as seen from the wrist cameras.}
    \label{fig:mosaic}
\end{figure}

\section{Related Work}
\label{sec:related_work}

\subsection{Data Collection Paradigms for Robotic Manipulation}
\label{sec:rw_data}
Robotic manipulation demonstrations are commonly collected through distinct approaches, each exposing a different trade-off between scalability, embodiment alignment, and data quality.
Leader-follower teleoperation~\cite{zhao2023learning,fu2024mobilealoha,wu2024gello,buamanee2024bi} provides accurate robot-executable trajectories, but requires a physical robot at every collection station and often demands operator training, limiting throughput.
Human-centric sources, such as retargeting from human motion~\cite{qin2023anyteleop,cheng2024television,fu2024humanplus} or learning from egocentric videos~\cite{grauman2022ego4d,punamiya2026egoverse,beingbeyond2025beingh0,ohkawa2023assemblyhands,lin2026simdex}, scale more easily and capture diverse everyday behavior.
However, the resulting observations are separated from robot action spaces by a substantial embodiment gap.
Large cross-embodiment datasets~\cite{o2024open,khazatsky2024droid} and synthetic data generation~\cite{mandlekar2023mimicgen} further expand data coverage, but fine-grained, contact-rich bimanual manipulation remains underrepresented.

Handheld gripper interfaces offer a complementary collection paradigm.
Since the operator directly manipulates the same end-effector used by the robot, these systems can acquire robot-compatible gripper poses and visual observations without operating a full robot during demonstration.
This work follows the handheld-interface direction, with an emphasis on scalable data collection for precise, two-handed manipulation.

\subsection{Hand-Held Gripper Interface}
\label{sec:rw_handheld}
The Universal Manipulation Interface (UMI)~\cite{chi2024universal} introduced a portable pistol-grip device to collect demonstrations at a low cost. 
Subsequent UMI-style systems improve collection scale and throughput~\cite{liu2025fastumi,zhaxizhuoma2025fastumi,lin2024data}, add tactile sensing for contact-rich manipulation~\cite{liu2025vitamin,li2025vitamin,liang2025alltact,xu2025exumi}, or replace SLAM-based wrist tracking with VR systems for reliable 6\,DoF gripper poses~\cite{zeng2025activeumi}.
Other robot-free approaches use hand motion capture~\cite{wang2024dexcap}, exoskeletons~\cite{fang2024airexo}, or AR-assisted capture~\cite{chen2024arcap}.

Recent UMI-style systems scale handheld collection to large corpora~\cite{liu2026rdt2,wei2026hifiumi,xu2026bimanual,wang2026xrzero}, use a smartphone for tracking and AR feedback~\cite{fang2026robopocket,patel2026bpp}, track from 360$^\circ$ cameras alone~\cite{yang2026kiwi}, or extend to whole-body control with egocentric head cameras~\cite{xu2026hommi}.
Others incorporate visuo-tactile sensing~\cite{wu2025freetacman} or additional modalities such as force/torque and audio~\cite{choi2026umift,cheng2026tacumi,wu2026tamen,hayes2026polyumi,luo2026omniumi}.
Wearable exoskeletons and gloves extend collection to dexterous robot hands~\cite{xu2025dexumi,fang2025dexop,zhu2026dexexo,xu2026realdexumi,yu2026seedumi,li2026aetherock,fang2025airexo2,lin2026artglove}, and Koala Gripper~\cite{hajjahmad2026koala} co-designs ergonomic finger linkages for collection and deployment.

Despite this progress, two bottlenecks remain important for large-scale bimanual data collection.
First, pistol-grip UMI devices place the operator's fingers away from the gripper pinch point, which reduces haptic transparency and makes fine fingertip control difficult.
Second, tracking presents a trade-off: SLAM-based methods can drift or fail under fast motion and weak visual texture~\cite{chi2024universal,lin2024data,liu2025vitamin}, whereas VR-based systems support reliable tracking but often require wearing the headset, adding neck fatigue during long sessions~\cite{zeng2025activeumi,xu2025exumi}.

YUBI is designed around these two bottlenecks: improving operator-side dexterity through a finger-aligned gripper, and supporting sustained collection through VR-based tracking.
These choices target scalable acquisition of precise bimanual demonstrations.

\section{Method}
YUBI is designed to curate massive, high-quality UMI-based data for bimanual dexterous manipulation tasks.
This section presents the design principles of YUBI (\Cref{sec:yubi}) and our setup with an integrated camera and VR rig, and a portable extension (\Cref{sec:op_setup}).
Full system and processing details are found in the supplement.

\subsection{YUBI Design}\label{sec:yubi}

The design of YUBI aims to shift away from the conventional pistol-grip mechanism used in prior UMI works~\cite{chi2024universal,liu2025fastumi,lin2024data} toward a yielding, finger-aligned actuation scheme.

\customparagraph{Limitations of prior UMI grippers}
Although existing UMI systems have demonstrated successful robot deployment using only handheld devices~\cite{chi2024universal}, they have critical limitations when applied to real-world scenarios at scale: limited adaptability to dexterous manipulation tasks and long-duration sessions due to their bulky, heavy interfaces.

The conventional pistol-grip interface introduces a mechanical offset between the operator's fingers and the gripper's pinch-point, significantly reducing haptic transparency~\cite{wu2025freetacman}. This lack of tactile feedback causes operators to apply excessive gripping force (overcompensation), while mechanical backlash in the gears further hinders fine motor control.

For the gripper tip design, prior works adopt Fin-Ray-type designs~\cite{chi2024universal,liang2025alltact,liu2025vitamin}, enabling soft, compliant grasping; however, their deformation characteristics result in poor positional repeatability and insufficient gripping force for heavy-object handling ($\ge$\,2\,kg)~\cite{xu20216dls,li2025vitamin}.
These limitations are particularly critical for real-world assembly tasks, which demand both precise handling of small components (\eg, nuts) and the ability to handle heavy industrial parts.

Lastly, the device weight induces early fatigue in the operator's wrist; $\approx$\,780\,g in the original UMI gripper and $>$900\,g when integrated with the VR controller~\cite{zeng2025activeumi,xu2025exumi}.
Such physical stress introduces trajectory noise and drift in demonstrations, undermining the dataset quality.

\customparagraph{Yielding finger-aligned design}
To address the above limitations, we introduce a yielding, finger-driven actuation design into YUBI.
Specifically, one gripper jaw is actuated by the thumb, while the opposing jaw is driven by the coordinated motion of the index and middle fingers. 
Each jaw yields directly to its driving finger, so the gripper aperture follows the operator's natural pinch motion without motor-driven resistance.
This configuration mitigates control mismatch and improves haptic transparency, enabling operators to directly leverage their inherent dexterity in manipulation tasks.

To preserve the precision of finger-driven actuation while targeting a design payload of up to roughly 2\,kg, YUBI employs a dimensionally stable gripper design.
An integrated support grip serves as a mechanical fulcrum; while the index and middle fingers actuate the jaws, the remaining fingers stabilize the grip to distribute loads across the entire hand. Additionally, the finger geometry is optimized to balance reach against load capacity. 
By minimizing the moment arm, YUBI ensures the structural stiffness required for heavy objects without sacrificing accessibility to confined spaces.

To mitigate wrist fatigue caused by previous heavy designs (780--900\,g), YUBI adopts a miniaturized gripper architecture and a lightweight camera module, reducing the handheld mass to approximately 319\,g (200\,g gripper + 119\,g VR controller).

Overall, YUBI's finger-aligned design resolves the trade-off between dexterous manipulation and heavy-object handling while reducing operator fatigue for sustained data collection.
We open-source the full YUBI gripper build with 3D printed parts at under \$200~USD per unit (excluding the Quest~3S tracking system).

\customparagraph{Robot-transferable motion}
Beyond ergonomics, the finger-aligned design also shapes operator motion to aid cross-robot transfer. 
Because the support grip sits below the jaws, the gripper naturally points downward with the hand above the contact point, encouraging overhead approaches rather than lateral sweeps near the table level. 
The resulting trajectories stay above the workspace, suppressing near-surface lateral motions that robot arms cannot reproduce without collision.
This yields demonstrations that transfer more safely across kinematically distinct embodiments.
A paired support-grip study is provided in \Cref{supp:physical_studies}.

\begin{figure*}[t]
\centering
\includegraphics[width=\linewidth]{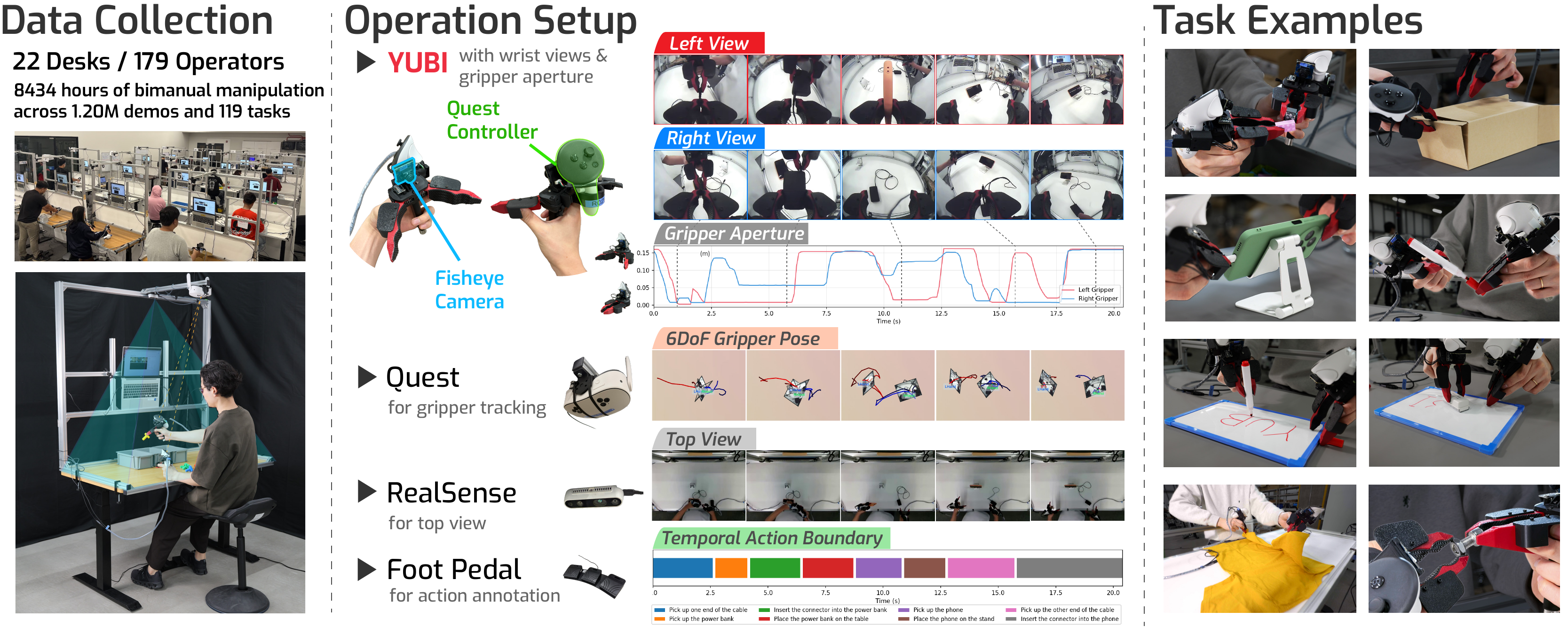}
\vspace{-3mm}
\caption{
\textbf{Overview:}
To ensure massive, high-quality robot manipulation data for scaling robot learning, we collect bimanual YUBI-based demonstrations at \Ndesk desks in parallel from \Nuser operators.
Our operation setups feature (i) a stereo top-view camera for stable workspace observation, (ii) a rig-mounted VR system for 6\,DoF gripper tracking where the VR controllers are attached to the YUBI gripper, and (iii) foot pedal-based action segmentation for hands-free control.
The collected dataset features diverse \DBNtask tasks that require various skills including assembling bricks and paper boxes, insertion, writing and erasing, folding clothes, and tightening nuts and bolts.
}
\label{fig:overview}
\end{figure*}

\subsection{Operation Setup}\label{sec:op_setup}
As shown in~\Cref{fig:overview}, we design an operation setup tailored to collect high-quality data where the operator grasps one YUBI device in each hand. 
For the sake of data quality, we adopt a fixed tabletop setup to operate in a stable environment as a primary option, while supporting a portable setup that brings the YUBI system beyond the tabletop environment. 
The stationary setup is integrated with VR systems (Quest) for 6\,DoF controller tracking, a top-view stereo camera (RealSense) for workspace monitoring, a task user interface (UI) on a laptop, and a foot pedal. 
These visual observations, trajectories, and gripper signals are transmitted to the laptop in real time.

\customparagraph{YUBI}
Two YUBI devices are installed on the desk, one for each hand.
Each YUBI device is equipped with an onboard wrist camera, a Quest controller, and a magnetic encoder to measure the gripper aperture, tracked at 100, 80, and 100\,Hz, respectively.
All streams are published as separate topics on a single ROS\,2 graph, each timestamped at the source under a shared clock.
Design details are found in \Cref{sec:yubi}.

\customparagraph{VR-based gripper tracking}
We use the Meta Quest 3S to track the 6\,DoF trajectory of the controller mounted on YUBI; unlike SLAM-based systems, which are prone to drift, scale ambiguity, and failure under fast motion or textureless environments~\cite{cadena2016past,murartal2015orbslam}, it does not depend on the scene's visual texture.
The Quest controllers are tracked by fusing infrared (IR) LED constellation observations from the HMD's inside-out cameras with the controller's onboard IMU. 
This active-marker + IMU pipeline performs robustly as in~\cite{holzwarth2021quest,carnevale2022quest}.
Unlike the head-worn VR-based UMI systems (\eg, ActiveUMI~\cite{zeng2025activeumi} and exUMI~\cite{xu2025exumi}), our setup mounts the heavy headset on the fixed rig.
This reduces neck fatigue while ensuring tracking coverage of the controllers from the mounted headset.
A quantitative comparison against OptiTrack ground truth is provided in \Cref{supp:tracking_accuracy}.

\customparagraph{Fixed stereo camera}
We use the RealSense D435, rigidly mounted on the forward-extending frame, to capture the task workspace.
This camera provides a stable top-down view of the workspace at 30\,Hz (RGB + depth).
The observations are used to identify low-quality demonstrations (\eg, mismatch between text instruction and action) and provide additional annotations via object detection, tracking, and VLM-based scene descriptions.
This stream is only used for data preprocessing.

\customparagraph{Portable YUBI for in-the-wild data collection}
While the rig-based setup is our primary configuration for stable data acquisition, the YUBI gripper itself is fully self-contained and can be operated in a portable mode that detaches from the desk rig (see \Cref{supp:portableyubi}).
This enables data acquisition in environments where a stationary rig is impractical, such as household tasks.
We attach the VR head-mounted display (HMD) to the chest and replace the top-view camera with an egocentric fisheye camera, rigidly integrated with the HMD. 
This enables collection in scenarios involving whole-body motion, \eg, ``put the tray in the dishwasher'', ``put the books away on the bookshelf'', ``hang the shirt on a hanger''.
The data schema (\Cref{sec:data_proc}) is preserved across modes, so portable and tabletop episodes share the same downstream pipeline.
We describe the details in the supplement.

\begin{figure}[t]
    \centering
    \begin{minipage}{0.52\linewidth}
        \centering        
        \includegraphics[width=\linewidth, trim=0 19.5bp 0 10.5bp, clip]{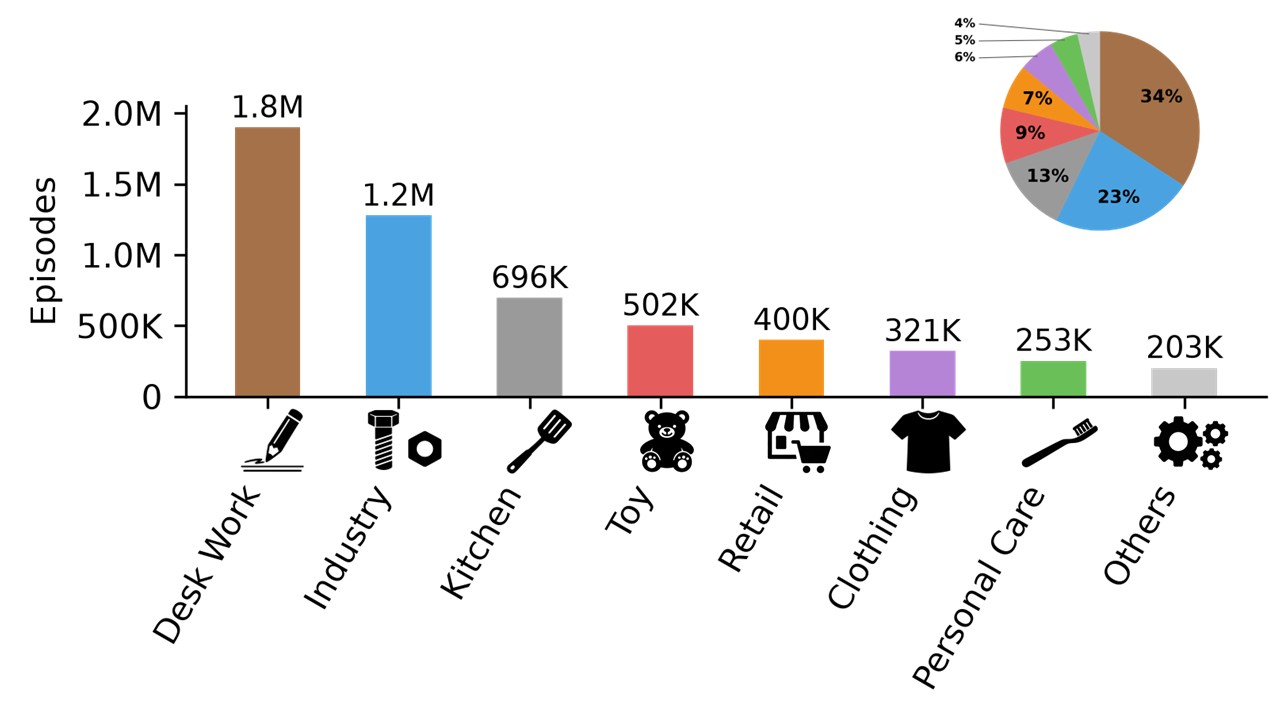}
        \vspace{-5.5mm}
        \caption{\textbf{Domain distribution:} The figure presents the share of the \DBep{} episodes across eight categories (desk work, industry, kitchen, toy, retail, clothing, personal care, and others), reflecting YUBI's target scope of precise, heavy, and everyday object handling.
        }
        \label{fig:task_domain}
    \end{minipage}
    \hfill
    \begin{minipage}{0.46\linewidth}
        \centering
        \includegraphics[width=\linewidth, trim=0 69bp 0 9bp, clip]{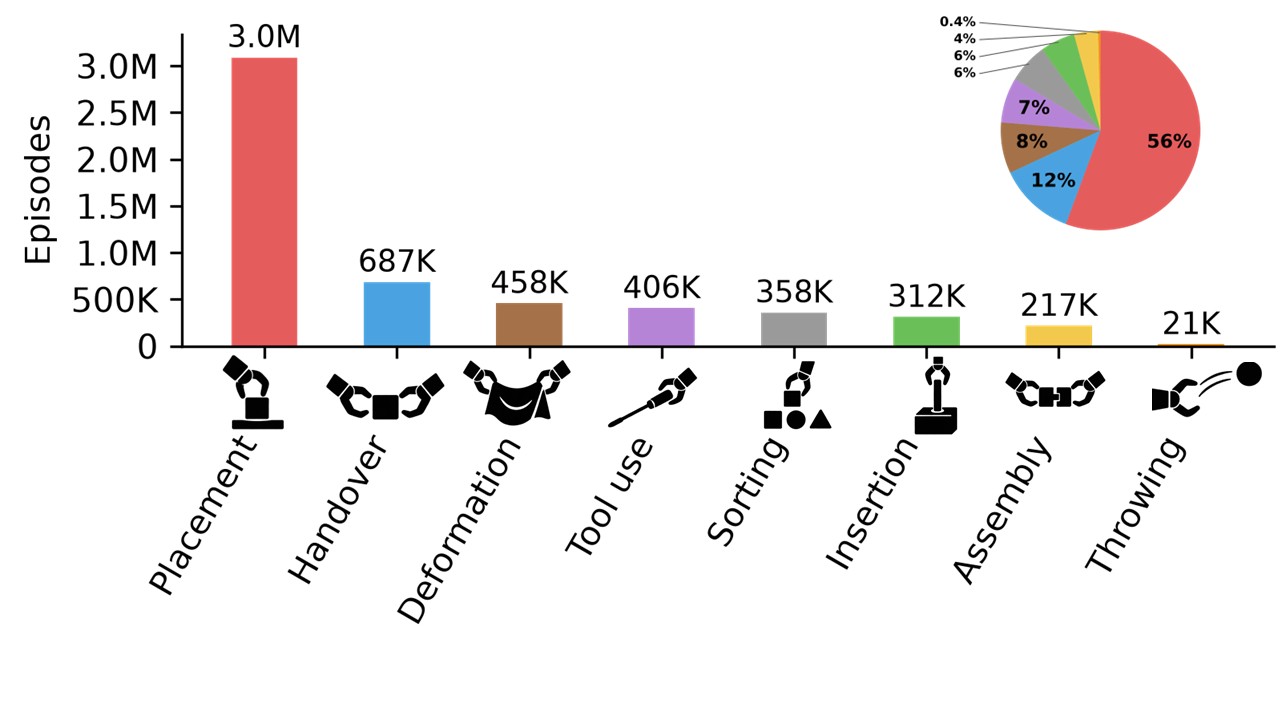}
        \vspace{-5mm}
        \caption{\textbf{Skill distribution:} The figure indicates the share of episodes by primary skill type, such as placement, insertion, assembly, sorting, deformation, and tool use. Labels reflect each episode's dominant skill; most tasks combine multiple skills in practice.}
        \label{fig:task_skill}
    \end{minipage}
\end{figure}

\section{Dataset}
\customparagraph{Schema}
Each recording is stored as a single episode file
that includes three synchronized 
\begin{wraptable}{r}{0.34\linewidth}
  \vspace{-1mm}
  \centering
  \caption{YUBI data statistics.}
  \label{table:data_stats}
  \vspace{1mm}
  \resizebox{0.92\linewidth}{!}{\begin{threeparttable}
    \begin{tabular}{lr}
    \toprule
    \textbf{Statistic} & \textbf{Value} \\
    \midrule
    \# Episodes                         & \DBep \\
    \# Demos\tnote{1}                   & \DBdemo\\
    Total duration (hrs)                & \DBhours \\
    \# Tasks                            & \DBNtask \\
Avg.\ sub-action / task             & \AvgSubTask \\
    Avg.\ episode duration (s)          & \AvgEpTime \\
    \bottomrule
    \end{tabular}
    \begin{tablenotes}
      \scriptsize
      \item[1] \hspace{-2mm} Total video-language-action pairs, \ie, the sum of sub-actions over all episodes.
    \end{tablenotes}
    \end{threeparttable}}
\end{wraptable} camera streams (two wrist cameras and RealSense), relative poses expressed in the wrist frames, and metadata for the task and sub-action.
Each task contains multiple primitive action instructions as shown in the action segment of \Cref{fig:overview} (avg.~\AvgSubTask actions per task).
Specifically, we store the translational components $\mathbf{t}^\text{right}, \mathbf{t}^\text{left} \in \mathbb{R}^3$ and rotational components $\mathbf{r}^\text{right}, \mathbf{r}^\text{left} \in \mathbb{R}^3$, where rotation is represented as Euler angles. 
We record the fingertip jaw angle for each hand, $d^\text{right}, d^\text{left} \in \mathbb{R}$.
All sensor streams are initially recorded in rosbag2 format at their native frequencies and are subsequently converted to the LeRobot format~\cite{cadene2024lerobot}, standardized at 30\,Hz.

\customparagraph{Statistics}
As shown in \Cref{table:data_stats}, we collected YUBI-based manipulation data at scale on \Ndesk desks.
The resulting data comprise \DBhours hours across \DBep episodes.
Data collection was conducted 24/7 over \DBperiod by \Nuser operators (\Nuserm male, \Nuserf female).
Our dataset is significantly larger than previous UMI-based datasets, such as the Fast-UMI data ($\approx$\,60 hours and 22 tasks)~\cite{zhaxizhuoma2025fastumi} and the original UMI data (12 hours and 4 tasks)~\cite{chi2024universal}.
It also features diverse domains and skills across \DBNtask tasks, as shown in \Cref{fig:task_domain,fig:task_skill}.

\begin{figure}[t]
    \centering
    \begin{minipage}{0.46\linewidth}
        \centering
        \includegraphics[width=\linewidth]{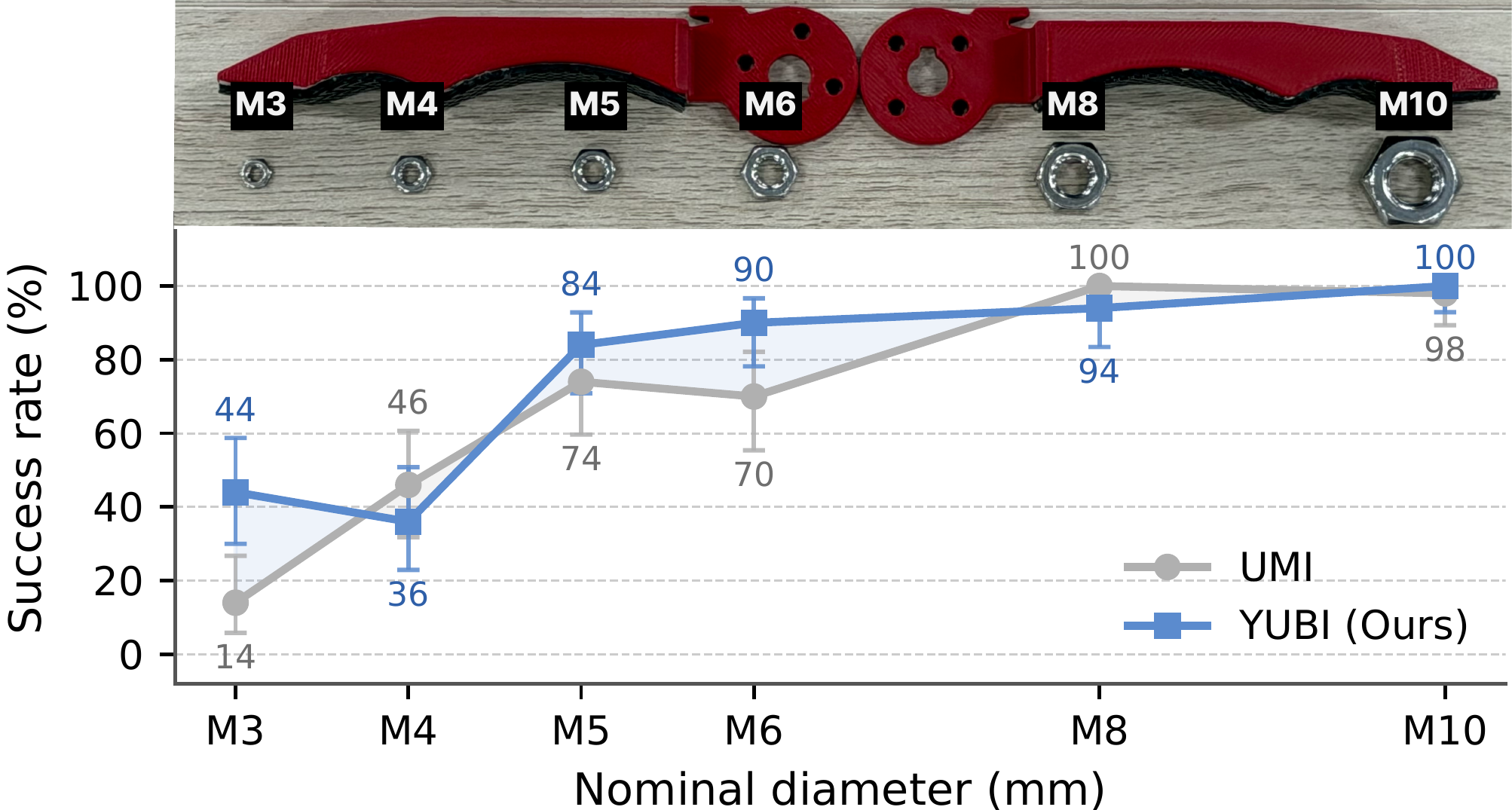}
        \vspace{-4mm}
        \caption{    
        \textbf{Dexterity test:} Pick-and-place success rate on hex nuts M10--M3 for UMI and YUBI.
Error bars show 95\% binomial confidence intervals (n=50).
}
        \label{fig:dexterity_test}
    \end{minipage}
    \hfill
    \begin{minipage}{0.52\linewidth}
        \centering
        \includegraphics[width=\linewidth]{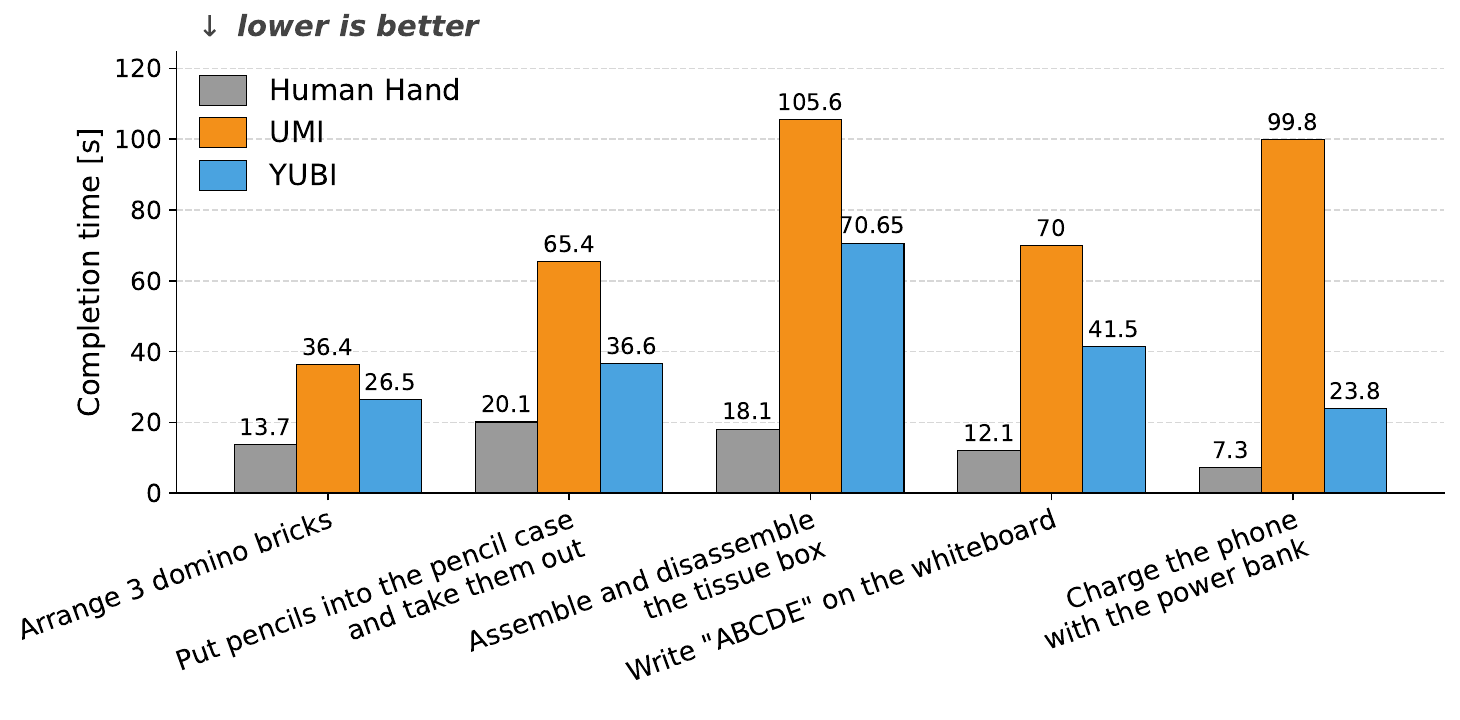}
        \vspace{-6mm}
        \caption{\textbf{Operation efficiency.} Mean completion time (seconds) on five tasks for Hand, UMI, and YUBI. YUBI is significantly faster than UMI and narrows the gap to direct hand operation.}
        \label{fig:efficiency_test}
    \end{minipage}
\end{figure}

\section{YUBI Usability Study}
We recruited a gender-balanced group of 10 operators who had no prior experience using either UMI or YUBI and ran the following user experiments.

\customparagraph{Dexterity test}
To assess the fine-grained dexterity of YUBI, the operators performed a single-attempt pick-and-place of six hex nuts (M10--M3, largest to smallest) from a table into a tray. As in \Cref{fig:dexterity_test}, both devices approach the ceiling on large nuts ($\ge 94\%$ at M8--M10) but diverge as the diameter shrinks: YUBI leads UMI by $+20$ and $+10$~pp at M6 and M5, and by roughly $3\times$ at the smallest M3 nut. These results indicate that YUBI is substantially more adaptable to precision tasks.

\customparagraph{Operational efficiency test}
To evaluate efficiency, the operators were instructed to perform five tasks under three conditions: direct hand operation (Hand), UMI~\cite{chi2024universal}, and YUBI, with counterbalanced ordering and five trials per condition. As shown in \Cref{fig:efficiency_test}, YUBI is consistently faster than UMI, with per-task speed-ups from $1.37\times$ (domino arrangement) to $4.19\times$ (phone charging), substantially narrowing the gap to direct hand operation even for precision tasks.

\section{Cross-Embodiment VLA Deployment}
\label{sec:deployment}
To evaluate whether the YUBI dataset translates into real-world robotic capability, we
train a VLA policy on YUBI's wrist data and deploy it across three bimanual robot platforms, UR~\cite{ur2026}, Franka~\cite{franka2026}, 
and Toyota ELEY~\cite{eley2026}, each equipped with the YUBI gripper as the end-effector.
Since the policies are trained on the gripper end-effector trajectory rather than in a robot-specific joint space, curated datasets can be reused across these robot arms without retargeting.
The deployment is only done with the wrist camera and trajectory data, not using the top-view observations.

\customparagraph{Policy training and inference}
As our primary policy, we adopt the vision-language-action model $\pi_{0.5}$~\cite{black2025pi05} trained jointly on multiple tasks in the end-effector action space.
Since the action space is the gripper trajectory rather than robot-specific joint commands, a single policy transfers across embodiments without retargeting. 
The policy is conditioned on a natural-language task instruction together with wrist RGB images at $30$\,Hz, and predicts relative end-effector poses in action chunks (\eg, $32$ steps). 
At inference, we execute the predicted end-effector trajectory at a downsampled control rate (\eg, $10$\,Hz) to keep commanded motion within the robot's velocity and acceleration limits, which are tighter than those of the human-collected demonstrations.
We also convert the commanded end-effector poses into joint-space targets via a per-robot inverse-kinematics (IK) solver.
Details are found in \Cref{sec:deploy_setup}.

\begin{table}[t]
  \caption{\textbf{Deployment results across embodiments.} Success rate (\%) of a $\pi_{0.5}$-based multi-task policy~\cite{black2025pi05} over $20$ rollouts per task. Gray brackets give 95\% Wilson confidence intervals.
}
  \vspace{-2mm}
  \centering
  \small
  \setlength{\tabcolsep}{3pt}
  \resizebox{\linewidth}{!}{\begin{tabular}{ccc ccc c}
  \toprule
  \multicolumn{3}{c}{\textbf{Bimanual UR}} & \multicolumn{3}{c}{\textbf{Bimanual Franka}} & \multicolumn{1}{c}{\textbf{ELEY}} \\
  \cmidrule(lr){1-3}\cmidrule(lr){4-6}\cmidrule(lr){7-7}
  Ball in basket & Stack cup pyramid & Unfold glasses & Place socks & Tape in box & Sort YUBI parts & Cup placement \\
  \midrule
  20/20 \textcolor{gray}{[84, 100]} & 13/20 \textcolor{gray}{[43, 82]} & 9/20 \textcolor{gray}{[26, 66]} & 16/20 \textcolor{gray}{[58, 92]} & 18/20 \textcolor{gray}{[70, 97]} & 2/20 \textcolor{gray}{[3, 30]} & 11/20 \textcolor{gray}{[34, 74]} \\
  \bottomrule
  \end{tabular}}
\label{table:deploy_embABC}
\end{table}
 \begin{figure*}[t]
    \centering    
    \includegraphics[width=\linewidth]{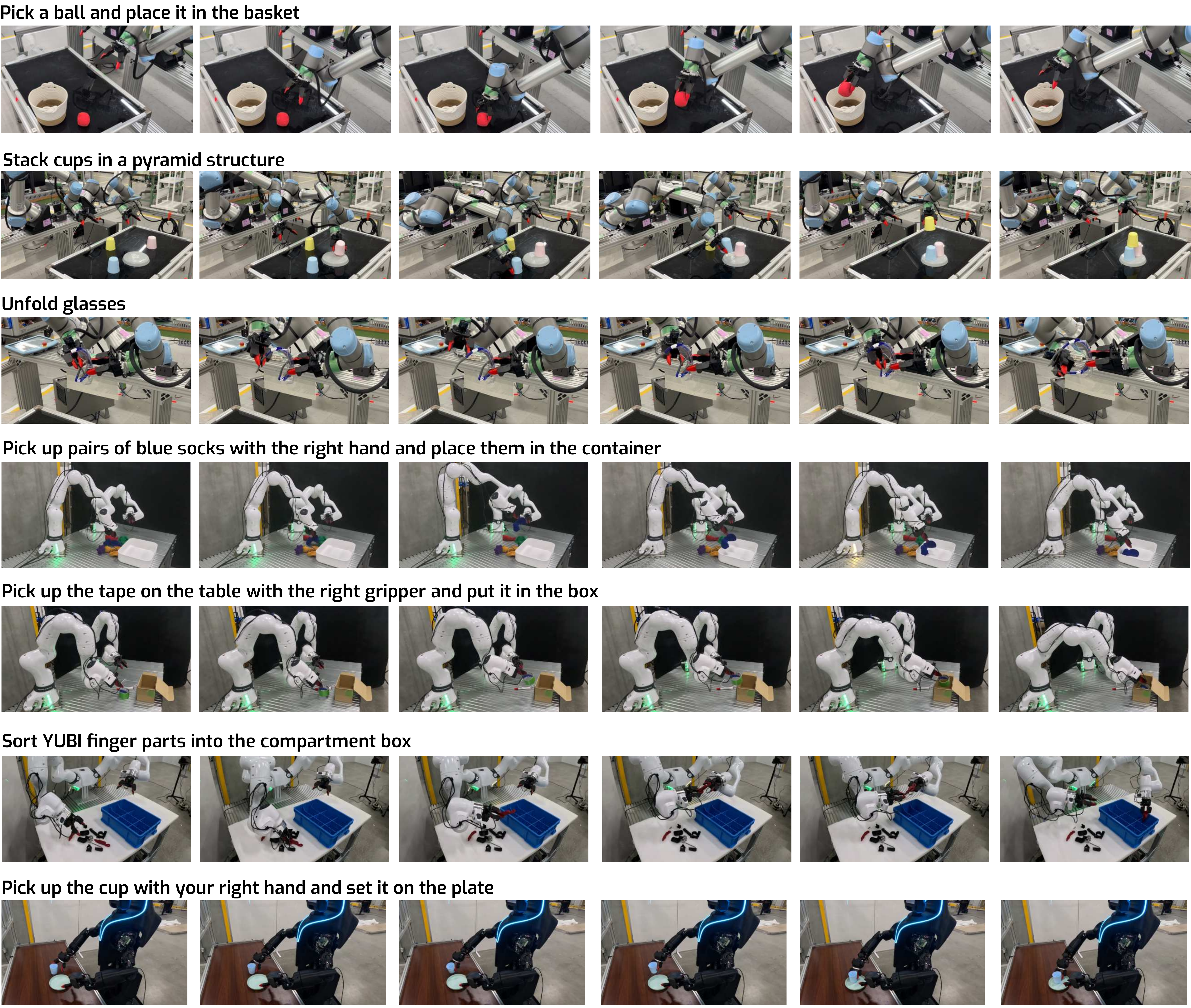}
\vspace{-4mm}
    \caption{\textbf{Robot deployment results.}
    We successfully deploy the policy trained on the YUBI data across three robots and six tasks.
    }
    \label{fig:robot_deploy}
\end{figure*}

\customparagraph{Results}
We execute 20 times per task and report the success rate in \Cref{table:deploy_embABC} (see \Cref{fig:robot_deploy} for qualitative examples).
Our preliminary study finds that the diffusion policy architecture~\cite{chi2023diffusionpolicy} proposed for the original UMI work~\cite{chi2024universal} converges quickly and performs well for a simple single-arm task, but struggles to generalize to complex bimanual tasks.
This motivates us to leverage the pretrained knowledge of $\pi_{0.5}$ in our YUBI training setup.
We observe that our multi-task VLA policy generalizes across various tasks, including distinct object types such as ball, cup, articulated glasses, differently colored socks, and tape.
We further demonstrate successful policy deployment on bimanual dexterity tasks,
such as ``stack cup pyramid'' and ``unfold glasses''. 
This confirms that YUBI's gripper-trajectory action space transfers across kinematically distinct robots.

\customparagraph{Large-scale pre-training}
To isolate the utility of an \DBhoursPT{}-hour pre-training corpus, we hold the backbone, loss, and action representation fixed. 
We run continual pre-training of $\pi_{0.5}$ on this \DBhoursPT{}-hour subset of the YUBI dataset for one epoch.
The pre-training only baseline barely solves the downstream tasks due to the broad action distribution of the corpus. 
To sharpen that distribution per task, we post-train from the pre-trained checkpoint on YUBI data (w/ pretrain), and without YUBI pre-training but starting with $\pi_{0.5}$ base checkpoint (w/o pretrain).

In 20 rollouts per task, post-training with pre-training achieves higher success, particularly on full-task success for sorting YUBI parts (10\%$\rightarrow$65\%) (\Cref{tab:pretrain}).
Our pretrained method achieves successful grasping of small parts, contact-rich manipulation, and bimanual handover prior to placement (see \Cref{fig:robot_deploy} for bimanual dexterity examples). 
This result provides evidence that 
(i) full-corpus pre-training is effective as an initialization for task-specific learning, 
and (ii) YUBI's trajectory and sensor alignment support learning this contact-rich bimanual skill.

\begin{table}[t]
  \centering
  \caption{Task-specific post-training on Franka with and without \DBhoursPT{}-hour pre-training. Entries report successes out of 20 rollouts with 95\% Wilson confidence intervals.
}
  \label{tab:pretrain}
  \small{
  \begin{tabular}{lccc}
    \toprule
    & Place socks & \multicolumn{2}{c}{Sort YUBI parts} \\
    \cmidrule(lr){3-4}
    Training & Full success & Grasp success & Full success \\
    \midrule
    w/ pretrain & \textbf{19/20} \textcolor{gray}{[76, 99]} & \textbf{19/20} \textcolor{gray}{[76, 99]}
& \textbf{13/20} \textcolor{gray}{[43, 82]}
\\
    w/o pretrain & 16/20 \textcolor{gray}{[58, 92]} & 13/20 \textcolor{gray}{[43, 82]} & 2/20 \textcolor{gray}{[3, 30]} \\
    \bottomrule
  \end{tabular}}
\end{table}

\section{Limitations and Discussion}
\label{sec:limitations}
While YUBI is a novel data collection tool, our study leaves several future directions open.
First, sub-millimeter precision and tactile-sensitive tasks, such as tight cable insertion and fragile material handling, remain challenging;
addressing this regime will require dedicated data curation, multi-modal sensing, and task-specific post-training.
\mbox{YUBI in the Dark}~\cite{shin2026yubiinthedark} shows that tactile sensing on hand-held grippers guides manipulation under limited vision, and \mbox{YUBI Go}~\cite{noguchi2026yubigo} extends YUBI to robot data collection anywhere with bimanual handheld grippers.
YUBI-STAG~\cite{tateno2026yubistag} explores contact-grounded interaction annotations for finer-grained language supervision to improve contact-rich manipulation.
Second, the optimal recipe for training VLA policies that combine YUBI data with complementary sources, \eg, in-the-wild demonstrations and real robot data, remains an open question for improving generalization.
To address variation in demonstration speed on YUBI data, RoboPace~\cite{shirasaka2026robopace} optimizes execution timing subject to physical and contact constraints.
Finally, our \DBhoursPT{}-hour pre-training study shows gains after task-specific post-training, but broader evaluation across tasks and further investigation of the scaling behavior and generalization of YUBI-trained VLAs are needed.
YUBI-Real2Sim~\cite{nishizawa2026yubireal2sim} explores scalable evaluation with reconstructed 3D Gaussian Splatting scenes to support reliable YUBI VLA development.

\section{Conclusion}
We present \textbf{Yielding Universal Bidigital Interface (YUBI)}, a novel finger-aligned gripper with yielding jaw actuation designed to collect massive, high-quality data for bimanual dexterous tasks.
The ergonomic gripper design enables fine control and heavy-object handling while reducing fatigue for sustained data collection.
Our operation setup also supports reliable gripper tracking with a decoupled VR system and stable visual observations from the top-view camera. 
The resulting dataset is the largest to date, comprising \DBhours hours of interaction data across \DBep episodes and \DBNtask distinct tasks. 
Our user study demonstrates the gripper's advantages in precise grasping and operational efficiency.
Furthermore, a multi-task policy trained on the YUBI dataset transfers across three bimanual robot platforms, validating that the collected data are directly executable as policy supervision across kinematically distinct robot arms paired with a common end-effector.
By open-sourcing the entire YUBI stack (hardware, software, and dataset), our work offers a scalable and reproducible path toward large-scale, high-fidelity data collection for robotic foundation models.

{\small
\bibliographystyle{plainnat}
\bibliography{refs/ref_umi,refs/ref_human2robot,refs/ref_vla,refs/ref_robots,refs/ref_umi_datasets}
}

\subsection*{Acknowledgments}
This paper is based on results obtained from a project, JPNP25015, commissioned by the New Energy and Industrial Technology Development Organization (NEDO).
The authors gratefully acknowledge the continued support of the AIRoA, Toyota Motor Corporation, and AIST research teams throughout this work.

\subsubsection*{Contributions}

\customparagraph{Project leadership} Takehiko Ohkawa and Jumpei Arima co-led the project.

\customparagraph{YUBI hardware} Yoshihiro Okumatsu, Yuki Wakayama, Daiki Fukunaga, and Yuki Noguchi designed the YUBI gripper.
Makoto Sugiura led the refinement and manufacturing of the YUBI hardware and co-designed the data-collection rig with Takehiko Ohkawa.

\customparagraph{YUBI application software} Jumpei Arima, Yuki Noguchi, Petr Khrapchenkov, Takuya Okubo, Naoaki Kanazawa, Naruya Kondo, and Miki Ma developed the YUBI data-collection software.

\customparagraph{Data infrastructure, curation, and annotation} Petr Khrapchenkov and Masafumi Takahashi built the data pipeline. Takuya Okubo, Masatoshi Tateno, and Hiroyuki Yamada led data analysis and curation. Masatoshi Tateno additionally led video production.

\customparagraph{Robot integration} Takuya Okubo, Tatsuya Kamijo, Yuki Noguchi, Jumpei Arima, and Makoto Sugiura integrated the YUBI gripper with robot platforms and built the policy deployment systems.

\customparagraph{VLA development} Yuki Noguchi and Takehiko Ohkawa led VLA post-training, while Takehiko Ohkawa led VLA pre-training. Kris Wu provided guidance on VLA model development.

\customparagraph{Policy evaluation} Yuki Noguchi, Hiroki Nishizawa, Kengo Ikeuchi, and Hayata Yuasa carried out policy evaluation.

\customparagraph{Operations and task definition} Takehiko Ohkawa managed operations and task definition with Yuma Shin, Jumpei Arima, Masatoshi Tateno, Kengo Ikeuchi, and Masaru Yajima. We additionally thank Kai Itatsu, Toki Go, and Yudai Tsuri for supporting overseas operations.

\customparagraph{Motion-capture evaluation} Kengo Ikeuchi and Takehiko Ohkawa led the motion capture-based evaluation of gripper pose tracking in cooperation with the AIST Digital Human Research Team, with special thanks to Tsubasa Maruyama, Mitsunori Tada, Ryosuke Hori, and Hiroyuki Deguchi.

\customparagraph{Field feedback} Koshi Makihara, Tomohiro Motoda, Floris Erich, and Yukiyasu Domae advised on industrial task definition and on YUBI development and deployment.

\customparagraph{Principal investigators} Kei Ota and Tatsuya Matsushima supervised the project as PIs.

\newpage
\appendix

\begin{center}
{\Large\bfseries Supplementary Material}\\[0.4em]
{\large for ``YUBI: Yielding Universal Bidigital Interface for\\
Bimanual Dexterous Manipulation at Scale''}
\end{center}
\vspace{0.6em}

\begin{figure*}[htbp]
    \centering
    \includegraphics[width=1\linewidth]{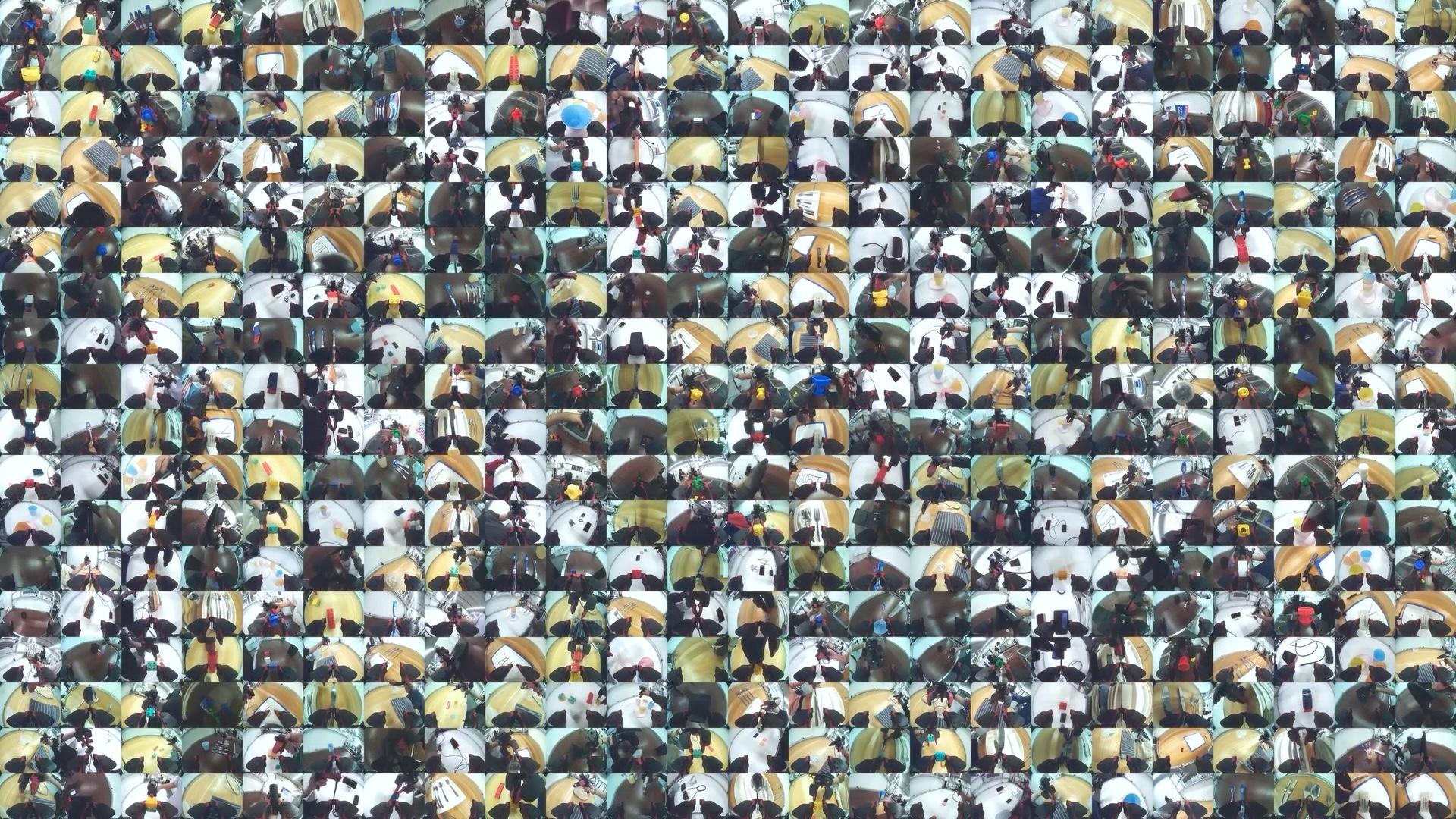}
    \vspace{-4mm}
    \caption{\textbf{Overview of collected YUBI data.}
    This figure shows wrist camera images, covering diverse distribution of the scene, tasks, and operators.
    }
    \label{fig:cover}
\end{figure*}

\section{Supplementary Video}
We provide supplementary videos that summarize the YUBI system, the data-collection process, and the experiments. 
The videos are presented in \href{https://yubi.airoa.io/}{https://yubi.airoa.io/}.

\section{Dataset Details}

\begin{figure*}[t]
    \centering
    \includegraphics[width=\linewidth]{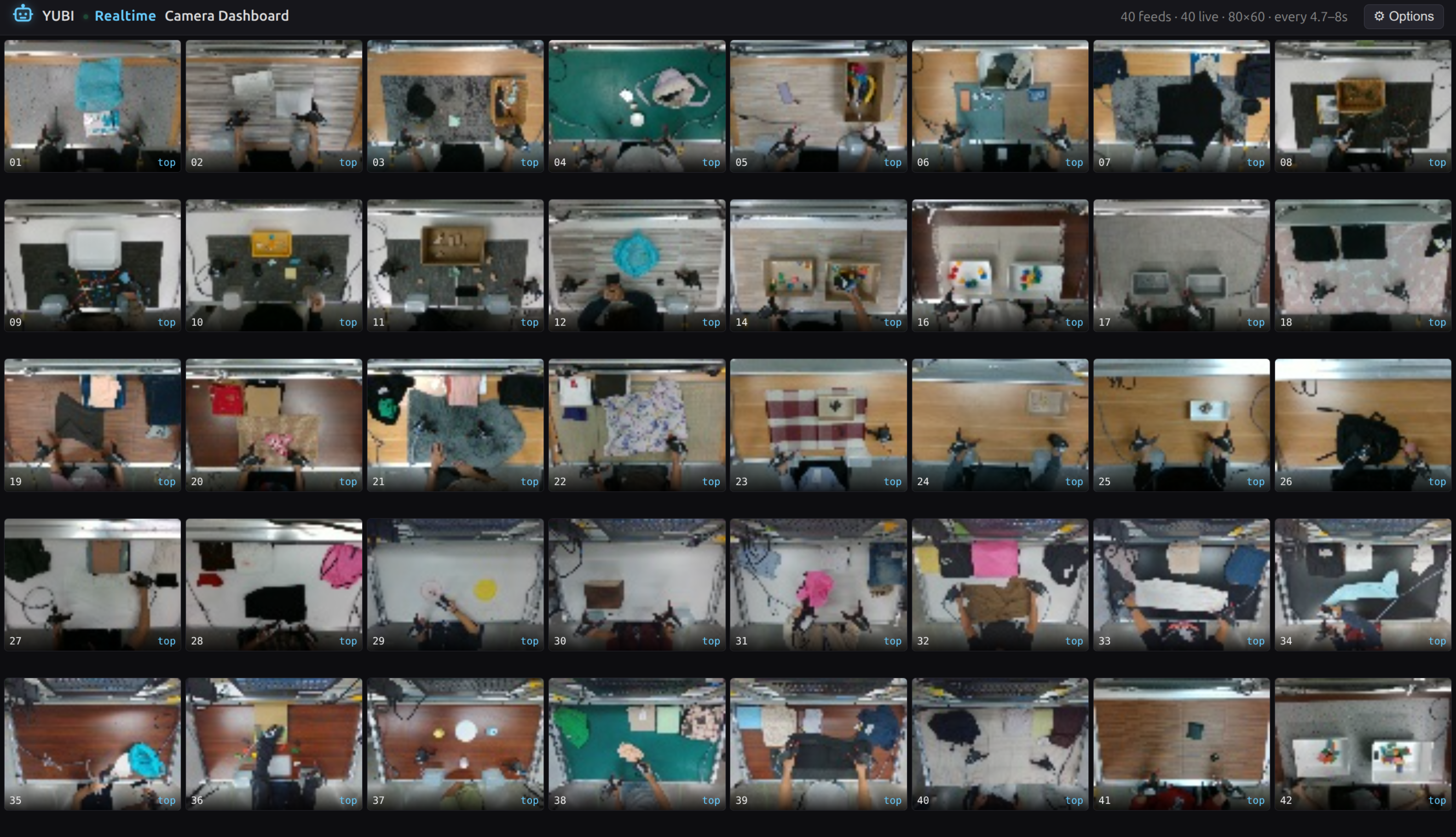}
\vspace{-4mm}
    \caption{\textbf{Live monitoring of the data-collection site.} A web-based dashboard aggregates real-time signals from every active operator station, including per-desk session status, camera previews, gripper-tracking quality, and cumulative episode counts. This allows operators and supervisors to monitor data-collection progress across all desks in parallel and to identify low-quality sessions or device failures as they occur.}
    \label{fig:dashboard}
\end{figure*}

\Cref{fig:cover} samples wrist-camera frames from across the \DBhours{} hours and \DBep{} episodes captured to date, spanning all \DBNtask{} tasks and \Nuser{} operators (\Nuserm{} male, \Nuserf{} female).
The diversity of scenes, objects, hands, and lighting visible in a single grid conveys the breadth of the dataset.
\Cref{fig:dashboard} then shows the operational backbone that enabled this scale, a live dashboard that simultaneously monitors all operator stations.
By surfacing per-desk session status, camera previews, gripper-tracking quality, and episode counts in real time, the dashboard turns multi-site data collection into a single supervisable process, enabling reliable quality assurance and sustained data collection throughput.

\begin{figure}[t]
    \centering
    \begin{minipage}{0.52\linewidth}
        \centering        
        \includegraphics[width=\linewidth]{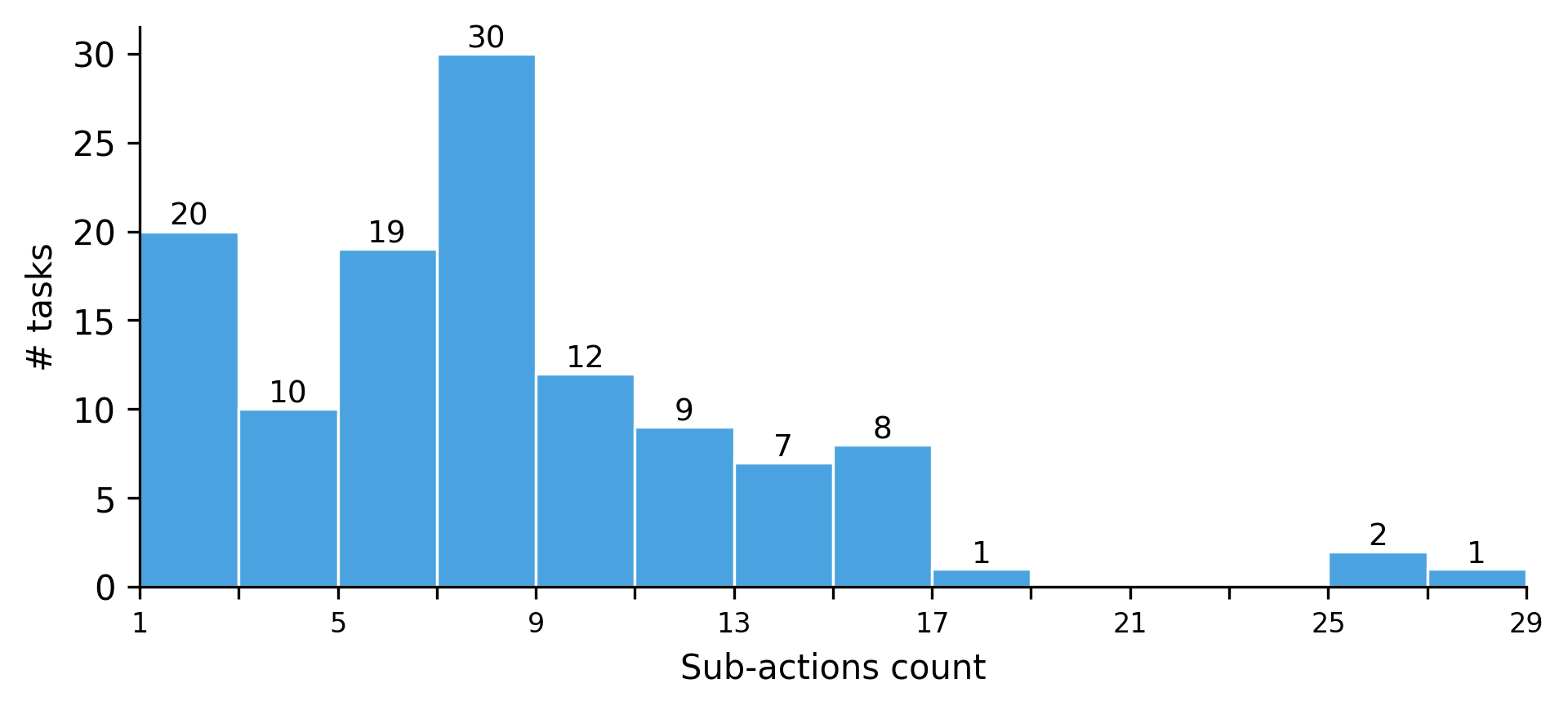}
        \vspace{-5.5mm}
        \caption{\textbf{Numbers of sub-actions for each task:} The figure presents the numbers of sub-actions for each task in a 119-task subset.
        }
        \label{fig:pa_count}
    \end{minipage}
    \hfill
    \begin{minipage}{0.46\linewidth}
        \centering
        \includegraphics[width=\linewidth]{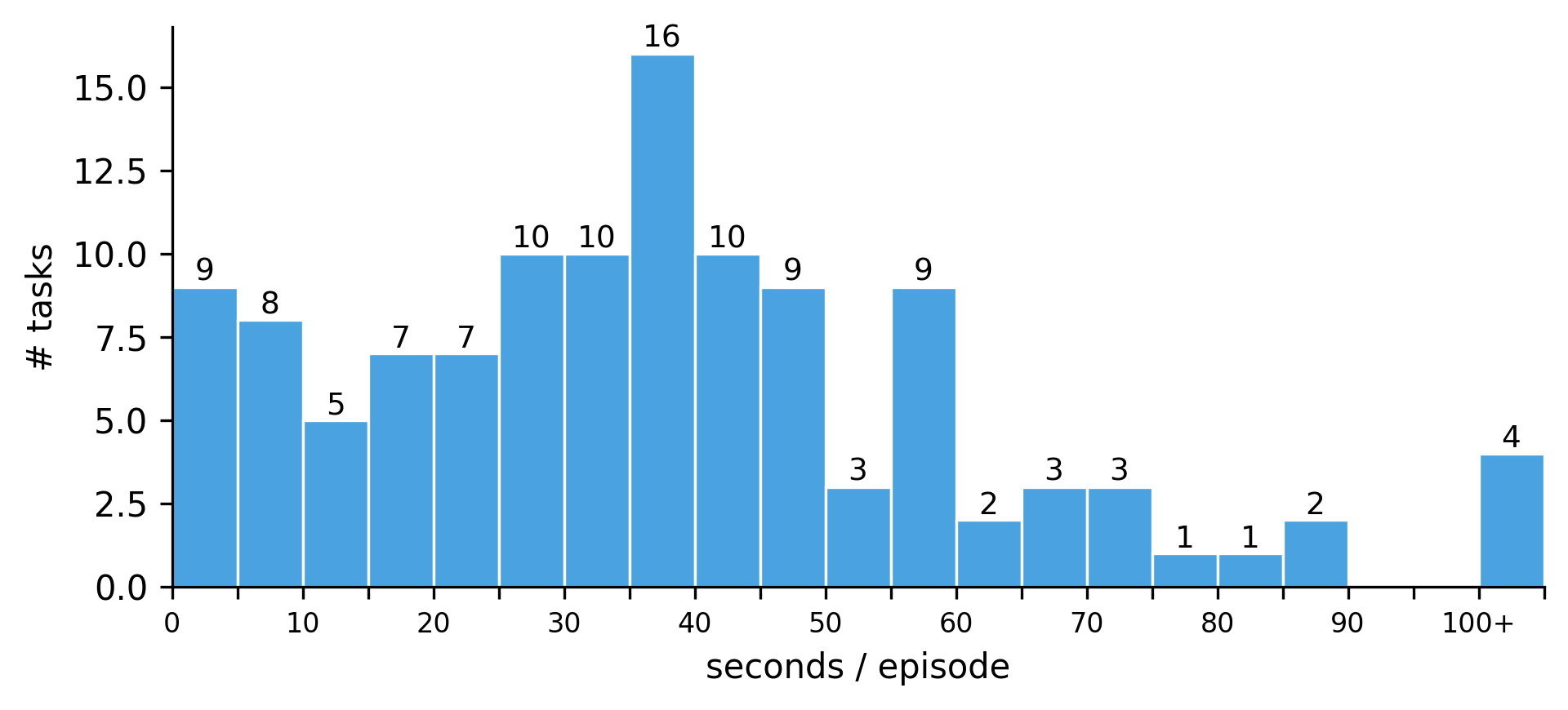}
        \vspace{-5mm}
        \caption{\textbf{Average episode seconds:} The figure indicates the average duration in seconds for completing each task in the same 119-task subset.}
        \label{fig:avg_episode_duration}
    \end{minipage}
\end{figure}

We further characterize the dataset along two task-level axes.
\Cref{fig:pa_count} plots the number of sub-actions per task across a 119-task subset.
Most tasks decompose into a handful of primitive sub-actions, while a long tail of compositional tasks (multi-step assembly, sorting of many parts, writing-and-erasing) extends well beyond ten sub-actions per recording.
\Cref{fig:avg_episode_duration} reports the average episode duration in seconds for each task.
Short pick-and-place demonstrations finish in less than a minute, whereas the longest tasks span further, in line with the sub-action distribution above.
Together the two figures show that the YUBI dataset spans a wide range of task complexity, from short atomic skills to long, compositional bimanual behaviors.

\subsection{Comparison with Prior UMI-based Datasets}
\begin{table}[t]
\centering
\caption{Comparison of UMI-style robot manipulation datasets.
\textbf{Track.}: gripper tracking method (SLAM-based vs.\ VR-based).
\textbf{Biman.}: \CIRCLE\ all tasks bimanual, \Circle\ all single-arm, \LEFTcircle\ mixed.
\textbf{Features}: distinctive sensing or embodiment beyond the standard Image+Proprio stream (\eg, Dex hand, Whole-body, Tactile, Audio).
\textbf{Dur.}: recording duration in hours.
}
\vspace{-2mm}
\label{tab:umi-data-comparison}
\renewcommand{\arraystretch}{1.2}
\setlength{\tabcolsep}{4pt}
\footnotesize
\begin{threeparttable}
\begin{tabularx}{\textwidth}{@{}l c c r c c c X@{}}
\toprule
\textbf{Project} & \textbf{Track.} & \textbf{\#Tasks} & \textbf{\#Demos} & \textbf{Dur.} & \textbf{\#Envs} & \textbf{Biman.} & \textbf{Features} \\
\midrule
\multicolumn{8}{@{}l}{\textit{Standard}} \\
\midrule
UMI (original) \cite{chi2024universal} & SLAM & 5 & 2.5K & 12 & 34 & \LEFTcircle & -- \\
LEGATO \cite{10855557} & SLAM & 6 & 0.9K & -- & 6 & \Circle & -- \\
FastUMI \cite{zhaxizhuoma2025fastumi} & SLAM & 22 & 9.0K & (60) & 22 & \Circle & -- \\
Data Scaling Laws \cite{lin2024data} & SLAM & 6 & 24.1K & -- & 160 & \Circle & -- \\
FastUMI-100K \cite{liu2025fastumi} & SLAM & 32 & 92.8K & 0.6K & 32 & \LEFTcircle & -- \\
\midrule
\multicolumn{8}{@{}l}{\textit{Multi-modality}} \\
\midrule
ManiWAV \cite{liu2024maniwav} & SLAM & 5 & 1.0K & -- & 11 & \Circle & Audio \\
exUMI \cite{xu2025exumi} & VR & 8 & 1.7K & 5 & 21 & \Circle & Tactile \\
Touch in the Wild \cite{zhu2026touch} & SLAM & 20 & 2.7K & -- & 20 & \Circle & Tactile \\
UMI-3D \cite{wang2026umi3dextendinguniversalmanipulation} & SLAM & 3 & 4.6K & -- & 8 & \Circle & LiDAR (point cloud) \\
\midrule
\multicolumn{8}{@{}l}{\textit{Dex}} \\
\midrule
DexUMI \cite{xu2025dexumi} & SLAM & 5 & 1.8K & -- & 5 & \Circle & Dex hand, F/T, Tactile \\
DexWild \cite{TaoT-RSS-25} & SLAM & 10 & 9.5K & 33 & 109 & \LEFTcircle & Dex hand \\
\midrule
\multicolumn{8}{@{}l}{\textit{Bimanual}} \\
\midrule
ActiveUMI \cite{zeng2025activeumi} & VR & 6 & -- & -- & -- & \CIRCLE & Head pose \\
HuMI \cite{nai2026humanoid} & SLAM & 5 & 0.9K & -- & 11 & \CIRCLE & Whole-body \\
GenRobot \cite{genrobot2026} & SLAM & 5 & 790K & 13K & 10K+ & \CIRCLE & Tactile \\
\midrule
\textbf{YUBI (Ours)} & VR & \DBNtask{} & \DBdemo{}\textsuperscript{\dag} & 54.6K & \Ndesk{} & \CIRCLE & Finger-aligned gripper \\
\bottomrule
\end{tabularx}
\begin{tablenotes}
  \scriptsize
  \item[\dag] We define a \emph{demo} as a single video-language-action pair. YUBI decomposes its \DBep{} recording episodes into \DBdemo{} demos via per-episode sub-action annotation.  
\end{tablenotes}
\end{threeparttable}
\end{table}
 
Table~\ref{tab:umi-data-comparison} situates YUBI among existing UMI-style manipulation datasets and clarifies our scaling philosophy. Against the largest UMI-style dataset reported in the literature, FastUMI-100K, YUBI provides about $41\times$ more tasks (\DBNtask{} vs.~$32$), $\sim$275$\times$ more demonstrations (\DBdemo{} vs.~$92.8$K), and $\sim$91$\times$ more recorded hours (\DBhours{} vs.~$600$). The gap widens further against earlier UMI-style releases: $\sim$2{,}700$\times$ more demonstrations than the largest tactile-, audio-, 3D-, or dexterous-hand-equipped entry (DexWild, $9.5$K).
The \emph{Multi-modality} and \emph{Dex} sections show scientifically valuable efforts whose scale remains proof-of-concept: every entry sits below $10$K demonstrations, and most stay below $3$K.
This is due to limited durability (\eg, fragile tactile pads), costly hardware and sensors, development cost for multi-fingered teleoperation hardware, etc. 

YUBI deliberately targets the orthogonal axis of \emph{operational scalability}: a lightweight, finger-aligned gripper that is inexpensive to fabricate, durable across thousands of collection hours, and intuitive enough that operators reach productive throughput with minimal onboarding. 
Beyond the scale, our dataset inherently features bimanual data collection essential for skillful tabletop tasks and adaptation to humanoid deployment.
The resulting dataset combines the breadth of real-world bimanual dexterous tasks with the scale needed to actually train robotic foundation models on them.

\begin{figure*}[htbp]
    \centering
    \includegraphics[width=0.6\linewidth]{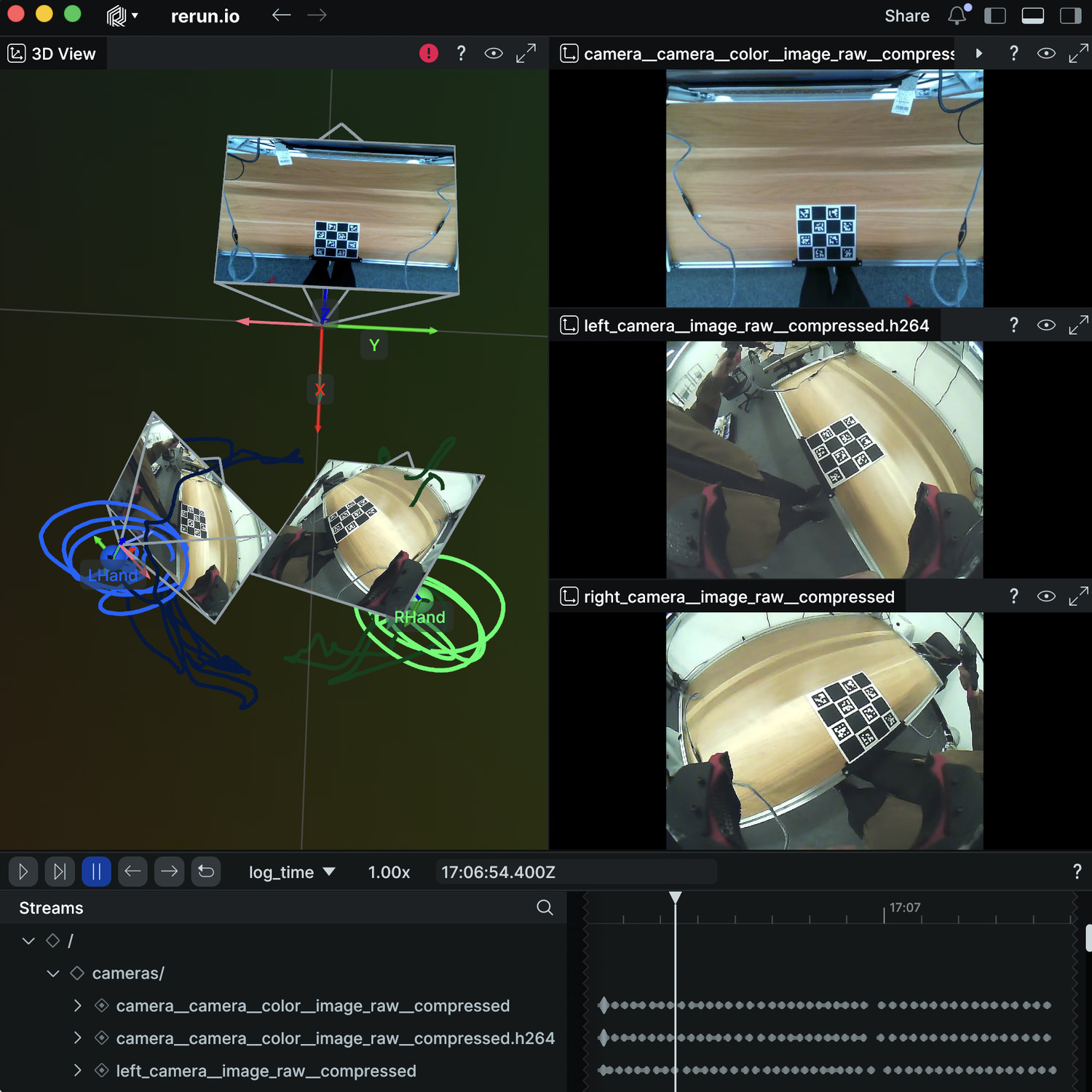}
\caption{\textbf{Extrinsic calibration process.}
    The ChArUco board is placed in the same position across all data collection desks and captured from all cameras to align the coordinates and reset the board to the world origin. 
    }
    \label{fig:calib}
\end{figure*}

\subsection{Data Processing}\label{sec:data_proc}

\customparagraph{Calibration}
Since the gripper trajectories have independent origins across different sessions or desk setups, we align them to a shared table frame via a dedicated calibration session, as shown in \Cref{fig:calib}.
We estimate a rigid transformation between the Quest and table coordinates, where a calibration board placed on the table is observed from multiple viewpoints using the wrist-mounted cameras of YUBI and the table-top camera.
We then compute the relative transformation between the calibration board (defining the table coordinate system) and the Quest tracking frame.
This transformation enables all tracked trajectories to be expressed consistently in the table-centered coordinate system.

\customparagraph{Filtering}
We remove defective episodes through a cascade of detectors as follows.
(i) We remove episodes shorter than a minimum-duration threshold, which can occur when operators accidentally press the foot pedal.
(ii) Stuck-signal detectors flag three freeze patterns:
\emph{full pose freeze}---a pose signal whose standard deviation is at or below numerical precision over the entire episode; \emph{partial pose freeze}---a contiguous run of frames whose per-frame translation $|\Delta\mathbf{t}|$ stays below a channel-specific noise floor for longer than a duration threshold; and \emph{aperture freeze}---a gripper whose jaw-angle variance is near zero across the episode.
(iii) Kinematic-plausibility detectors additionally drop episodes containing a single-frame translational jump exceeding an implausible velocity threshold given the recording rate, or a rotation increment greater than $90^\circ$ within a single frame.
(iv) Quest controller tracking can briefly become unreliable when the IR LEDs on a controller are occluded from the rig-mounted HMD (\eg, when the wrist is bent inward).
In our tabletop setup such occlusion is rare and short-lived.
To keep these segments out of training, we read the per-frame tracking-quality flag that Quest publishes alongside the controller pose to the ROS\,2 graph and drop any episode in which the flag indicates a degraded or untracked state for more than a small fraction of frames.
This filters episodes based on Quest's tracking-state flag; errors that occur while the controller is reported as tracked can remain undetected.

\subsection{Action-Segmentation Examples}
\label{sec:action_seg_examples}
\begin{figure}[htbp]
    \centering
    \includegraphics[width=\linewidth]{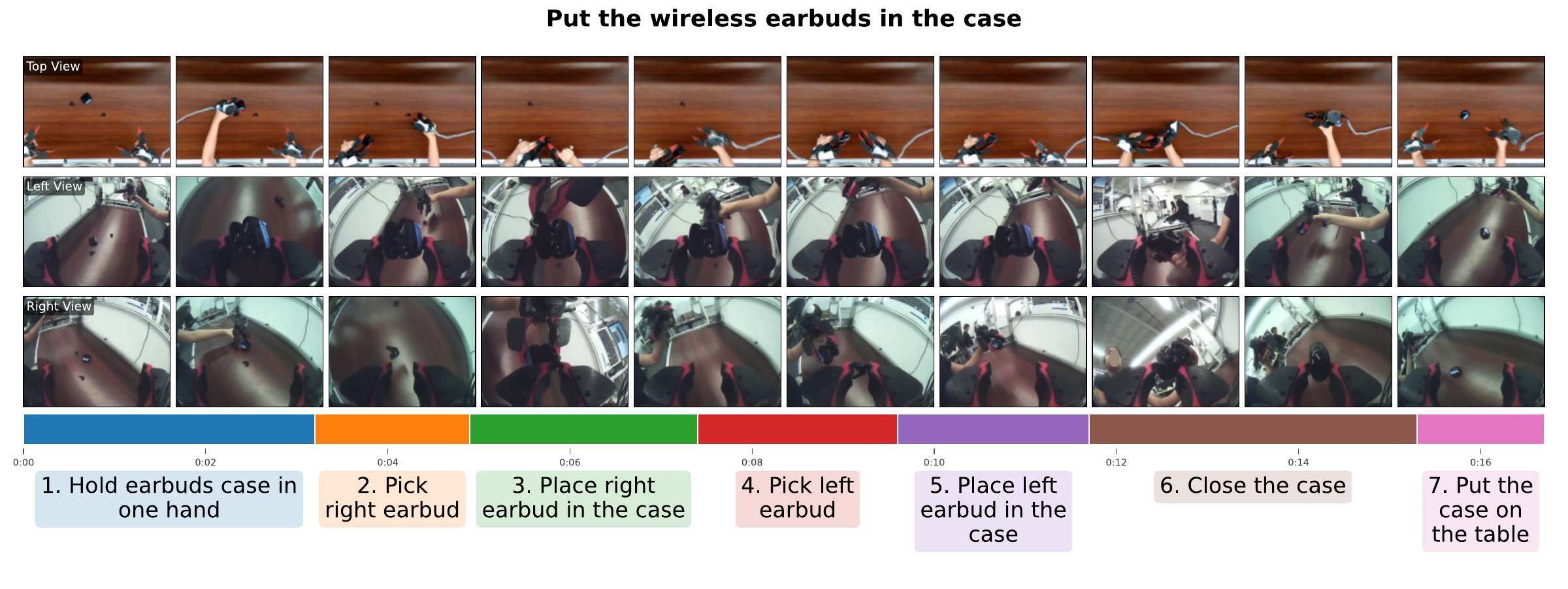}\\[3pt]
    \includegraphics[width=\linewidth]{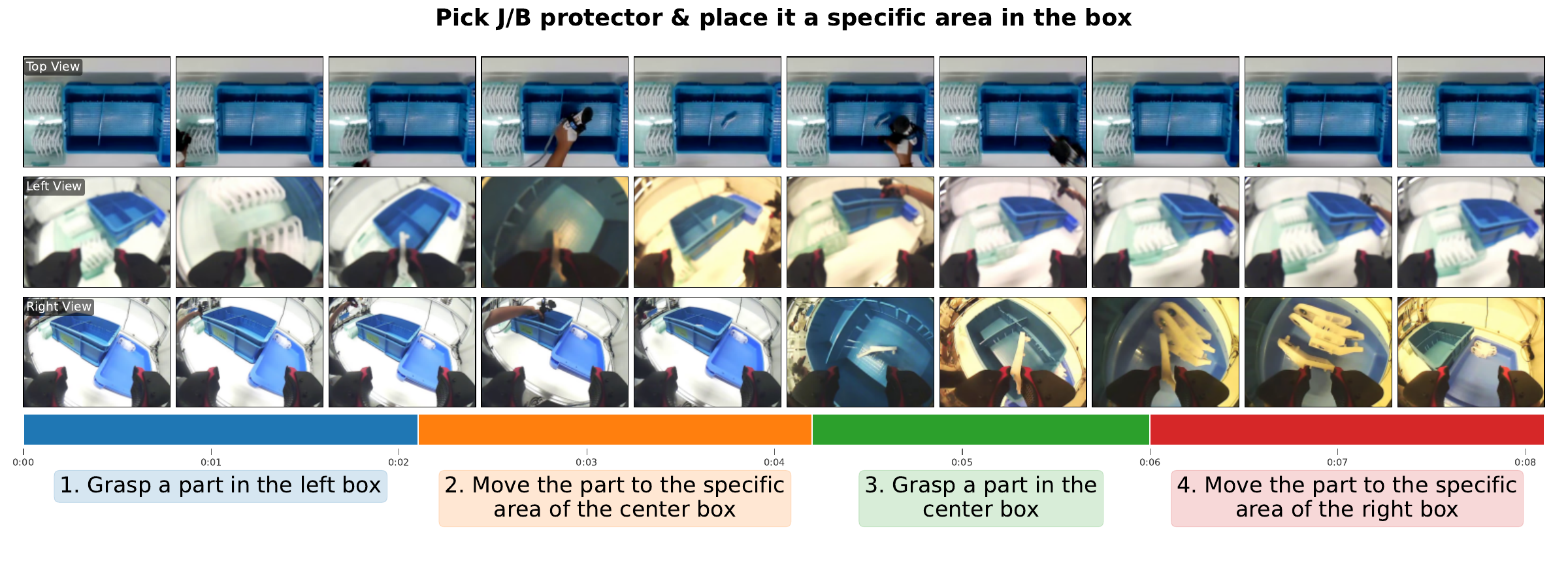}\\[3pt]
    \includegraphics[width=\linewidth]{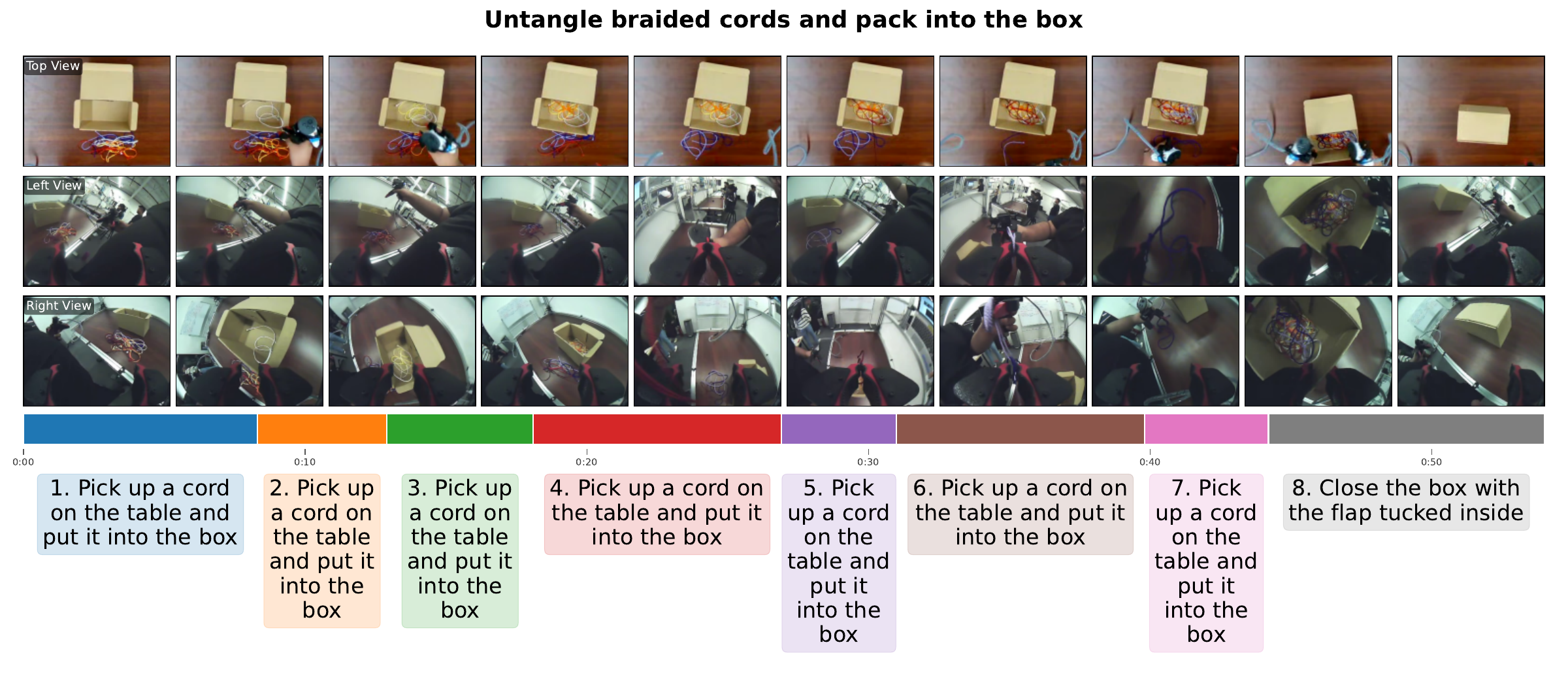}\\
    \caption{\textbf{Action segmentation examples (1/3).} Three representative YUBI episodes, each visualized as a timeline of segmented sub-actions with wrist-camera snapshots of the corresponding task progress.}
    \label{fig:action_seg_examples}
\end{figure}

\begin{figure}[htbp]
    \centering
    \includegraphics[width=\linewidth]{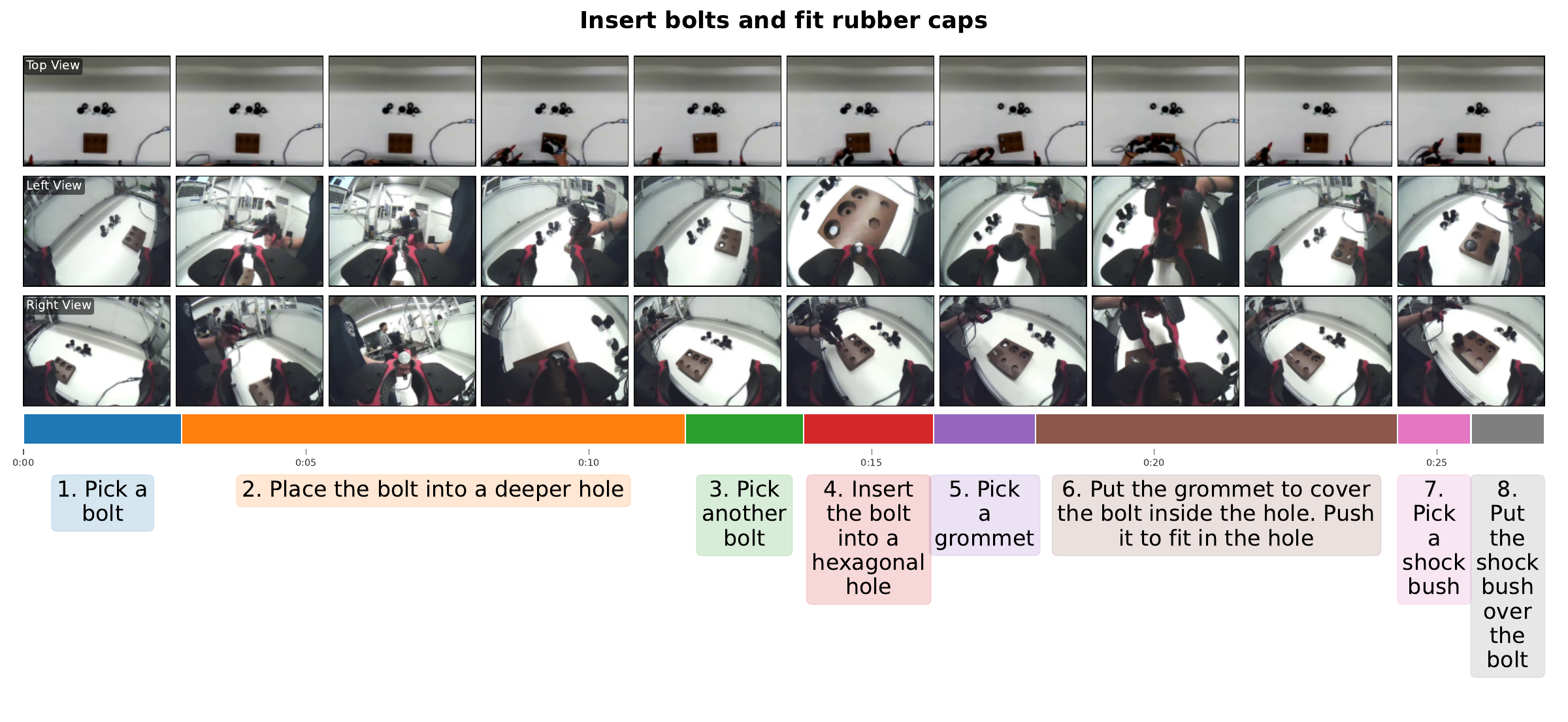}\\[3pt]
    \includegraphics[width=\linewidth]{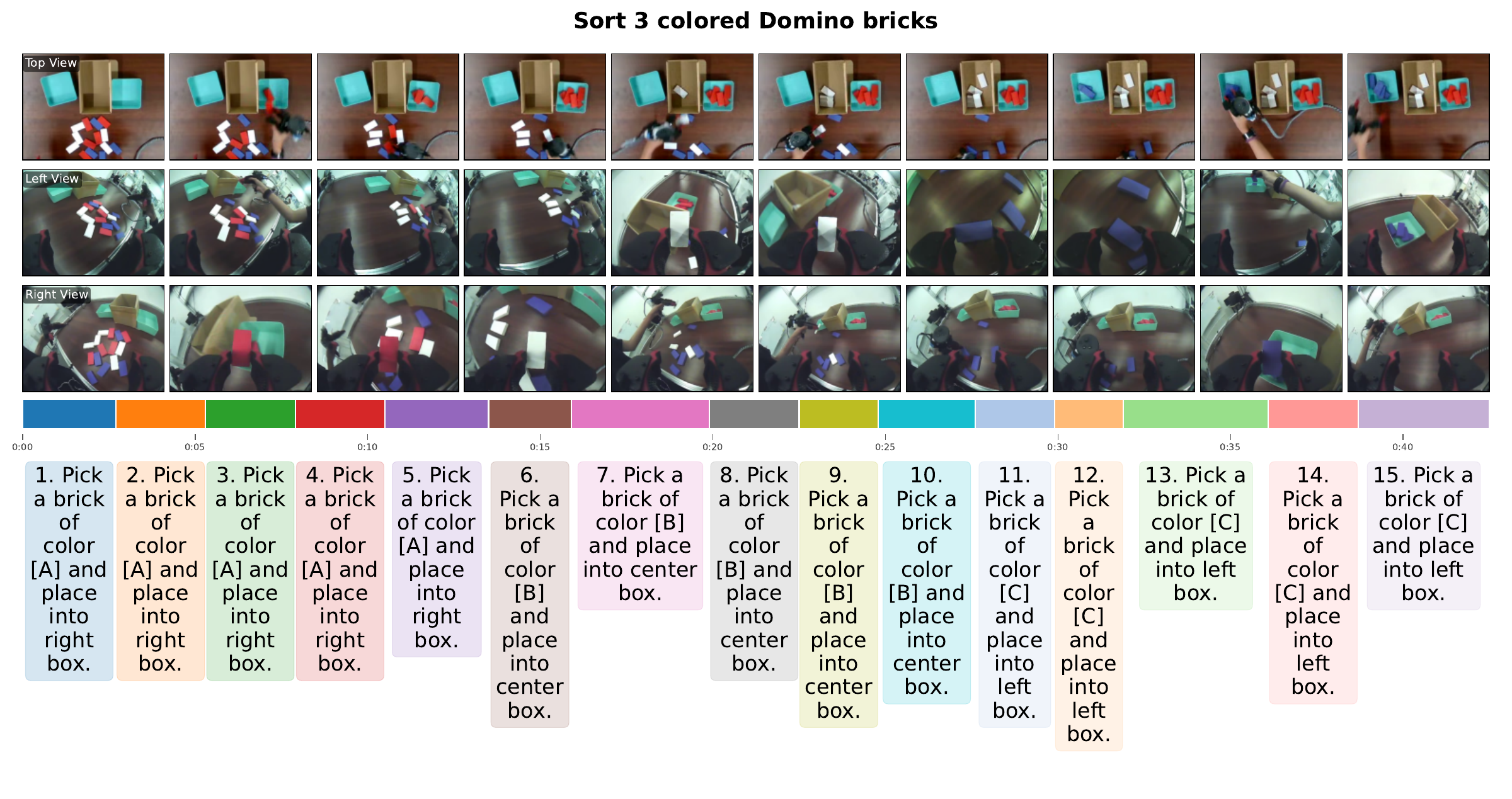}\\[3pt]
    \includegraphics[width=\linewidth]{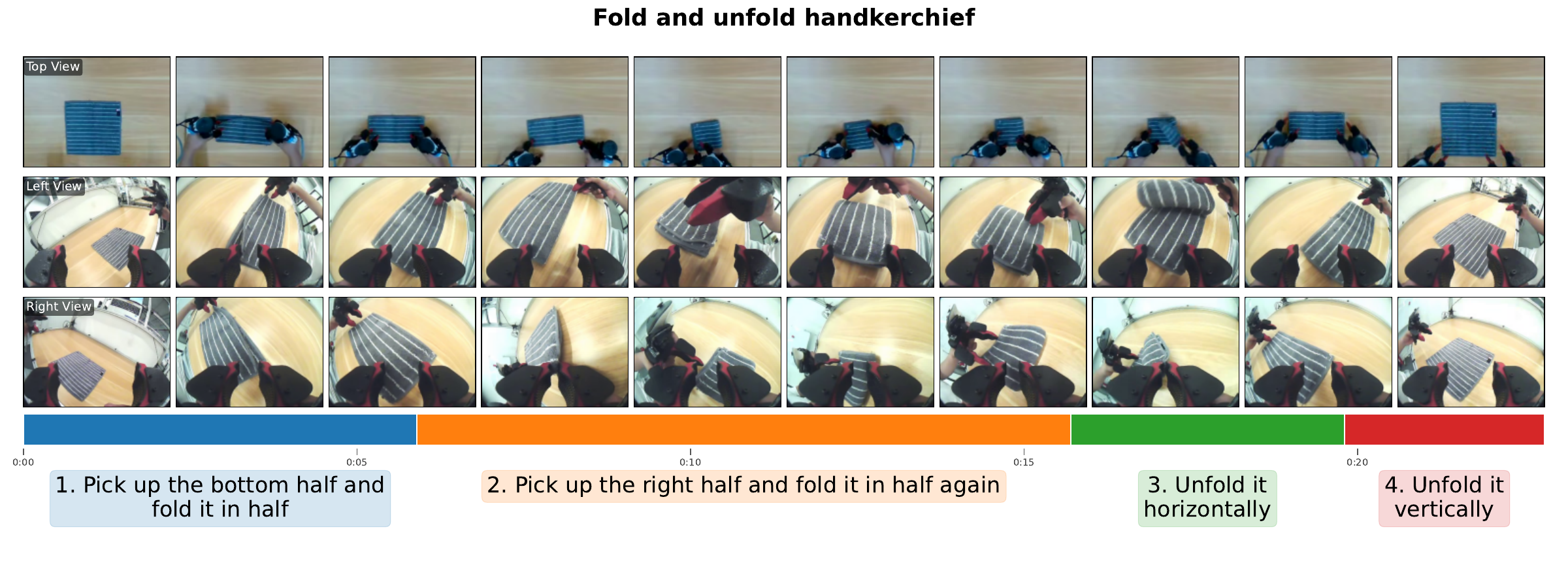}
    \caption{\textbf{Action segmentation examples (2/3).} Continued from \Cref{fig:action_seg_examples}.}
    \label{fig:action_seg_examples_b}
\end{figure}

\begin{figure}[htbp]
    \centering
\includegraphics[width=\linewidth]{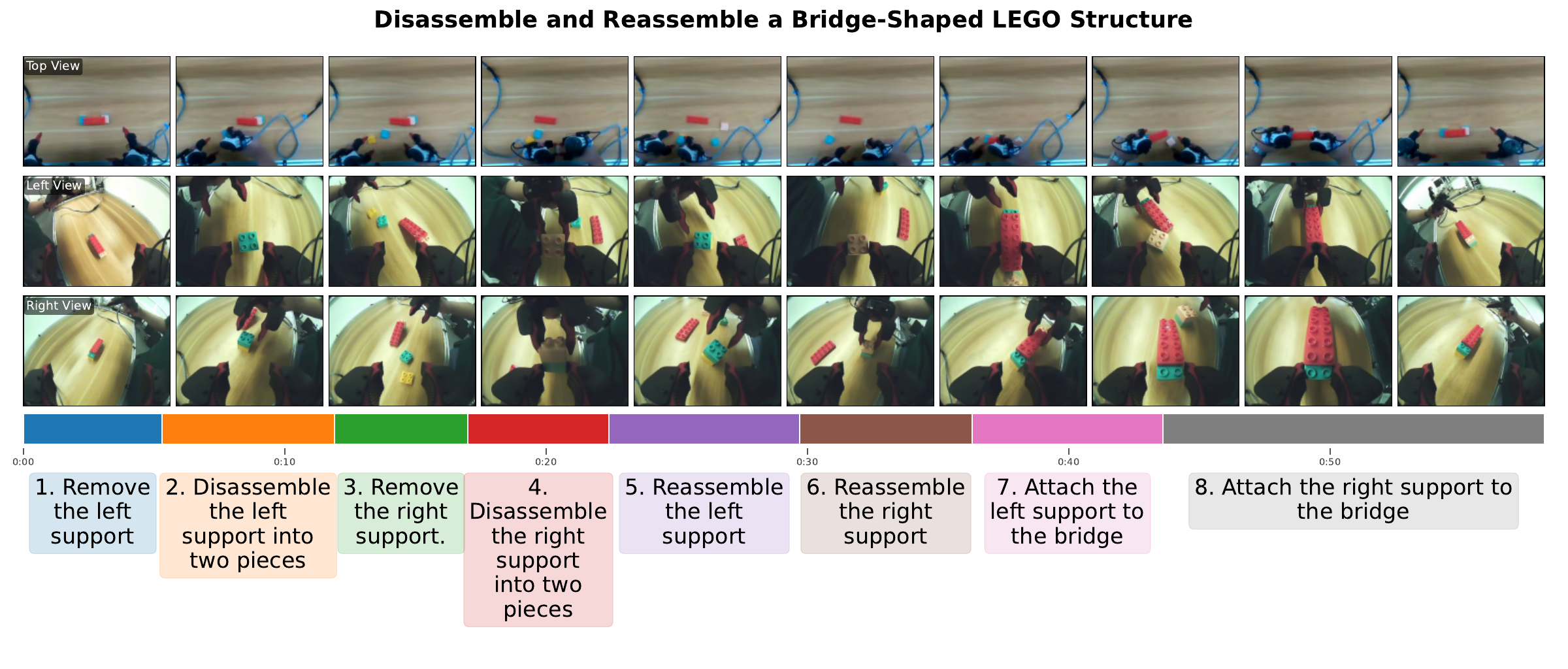}\\[3pt]
    \includegraphics[width=\linewidth]{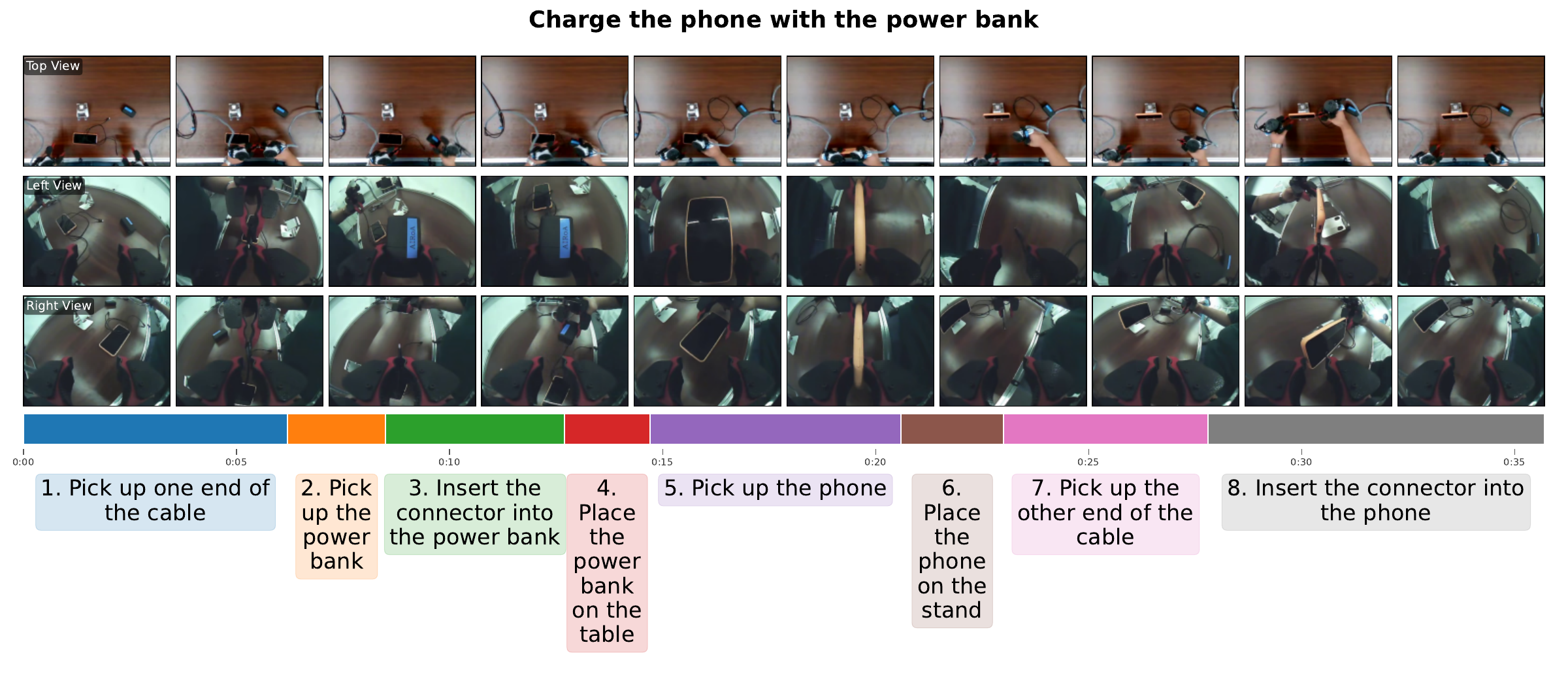}\\[3pt]
    \includegraphics[width=\linewidth]{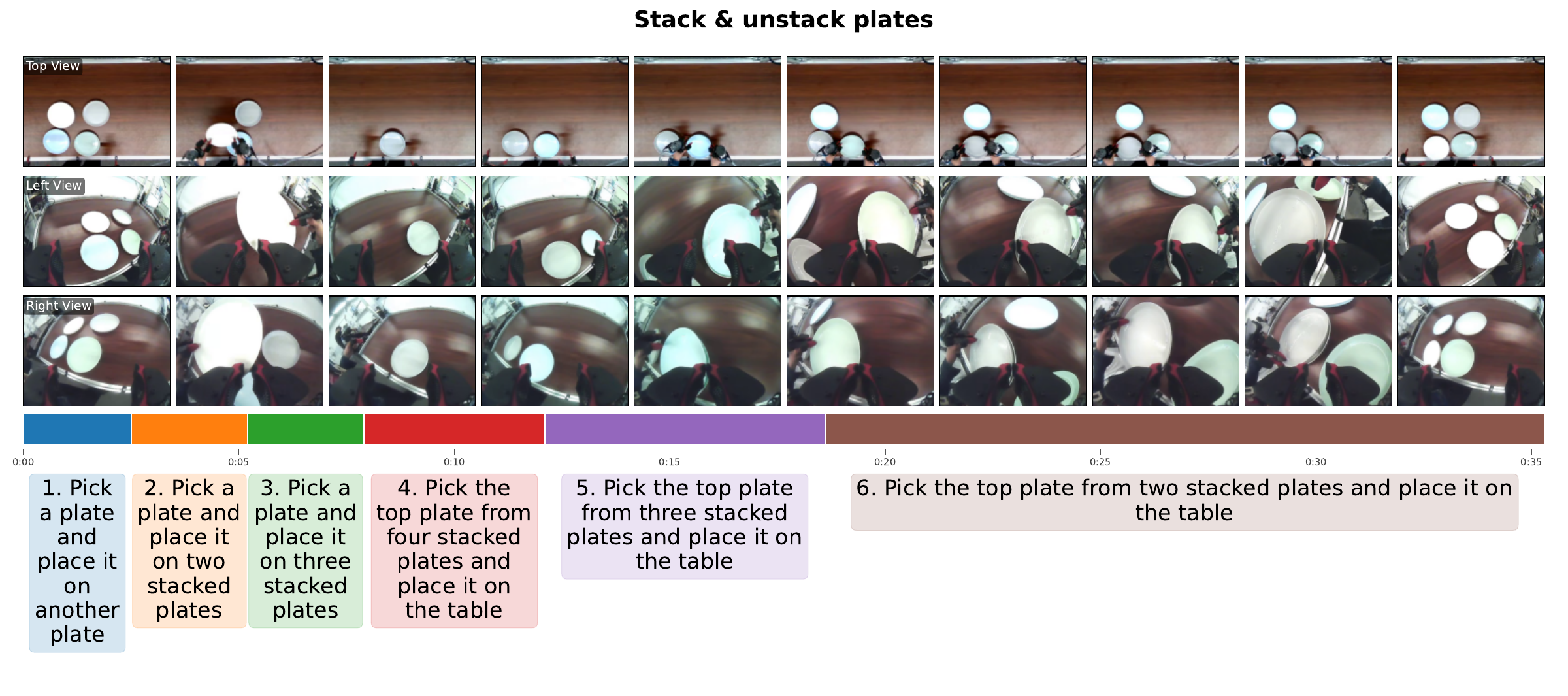}
    \caption{\textbf{Action segmentation examples (3/3).} Continued from \Cref{fig:action_seg_examples_b}.}
    \label{fig:action_seg_examples_c}
\end{figure}

Each YUBI episode is segmented into multiple primitive action steps that capture the sub-tasks an operator performs within a single recording. The action boundaries are inserted hands-free by foot-pedal events during collection (\Cref{sec:op_setup}) and refined in post-processing.
\Cref{fig:action_seg_examples,fig:action_seg_examples_b,fig:action_seg_examples_c} show representative episodes drawn from the \DBhours~hours of YUBI data. Each panel visualizes the wrist-camera observations together with the textual sub-action labels and their temporal extent on a shared timeline, illustrating the diversity of task structures captured by the dataset, from short pick-and-place sequences to longer assembly and writing tasks composed of several primitive sub-actions per episode.

\newpage

\newpage

\newpage

\clearpage

\section{Operation-Setup Details}

\subsection{Stationary Rig and Task UI}
\begin{figure}[t]
\centering
\includegraphics[width=\hsize]{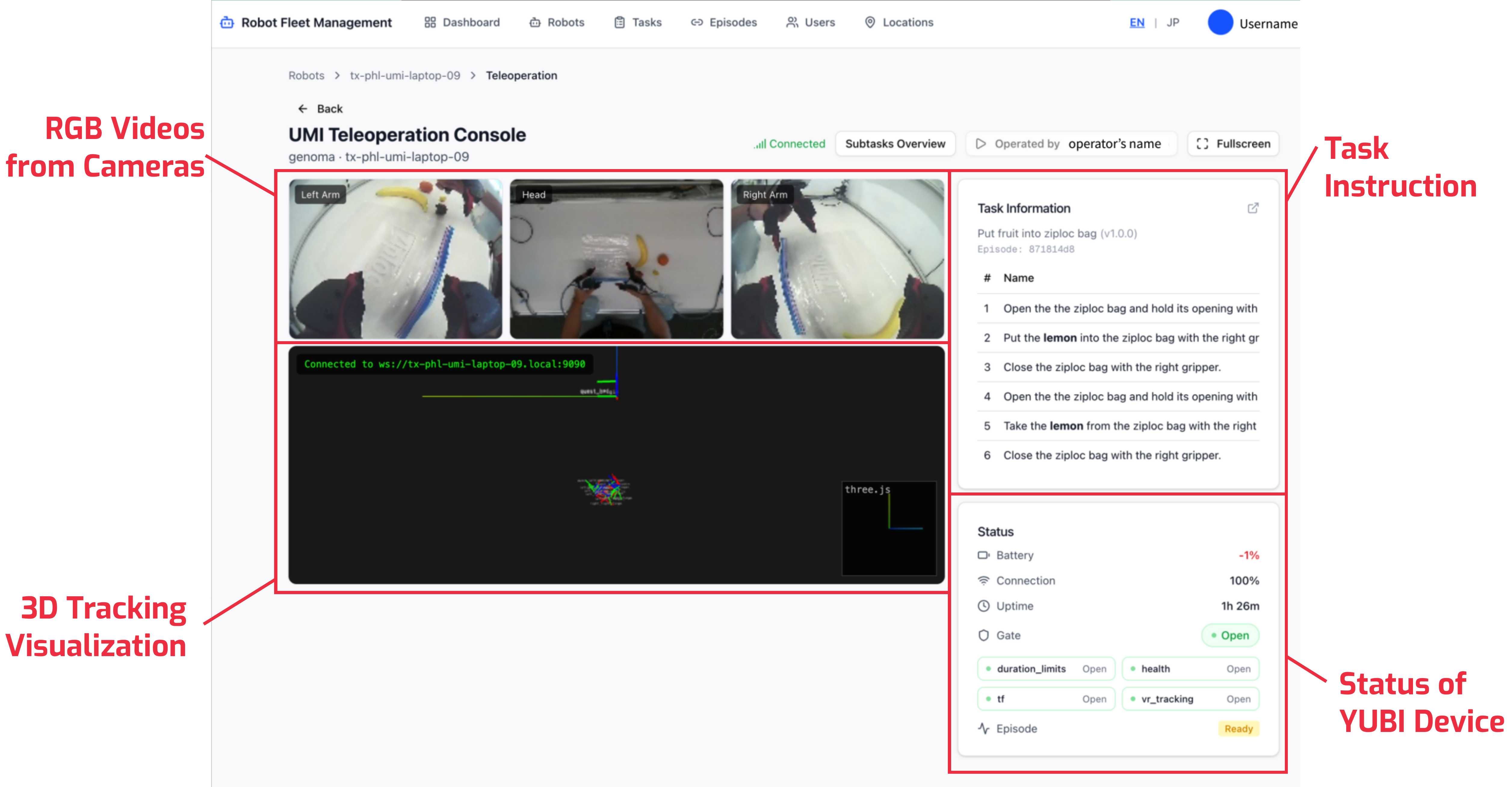}
\vspace{-2mm}
\captionof{figure}{\textbf{Task UI.} A rig-mounted laptop displays the current observations, 3D trajectory information, task and sub-action instructions, and per-sensor health status.
}
\label{fig:task_ui}
\end{figure}

We provide additional details for the operation setup found in \Cref{sec:op_setup}.
A laptop is mounted at the center of the frame wall, serving as the central hub of the system and the user interface for task instruction, as shown in \Cref{fig:task_ui}.
It aggregates all sensor streams while displaying real-time observations, tracking signals, and task information to the operator.
A major challenge in bimanual operation is that both hands are occupied by the grippers, leaving no free hand to annotate task transitions and sub-action boundaries through conventional inputs (\eg, buttons, touchscreens, keyboards).
To address this, we install a \emph{foot pedal} under the desk to start/stop action segments and to accept an episode for storage.
Operators press the pedal to indicate boundaries between predefined sub-action steps.

\subsection{Portable Configuration}\label{supp:portableyubi}

\begin{figure}[t]
    \centering
    \includegraphics[width=1\linewidth]{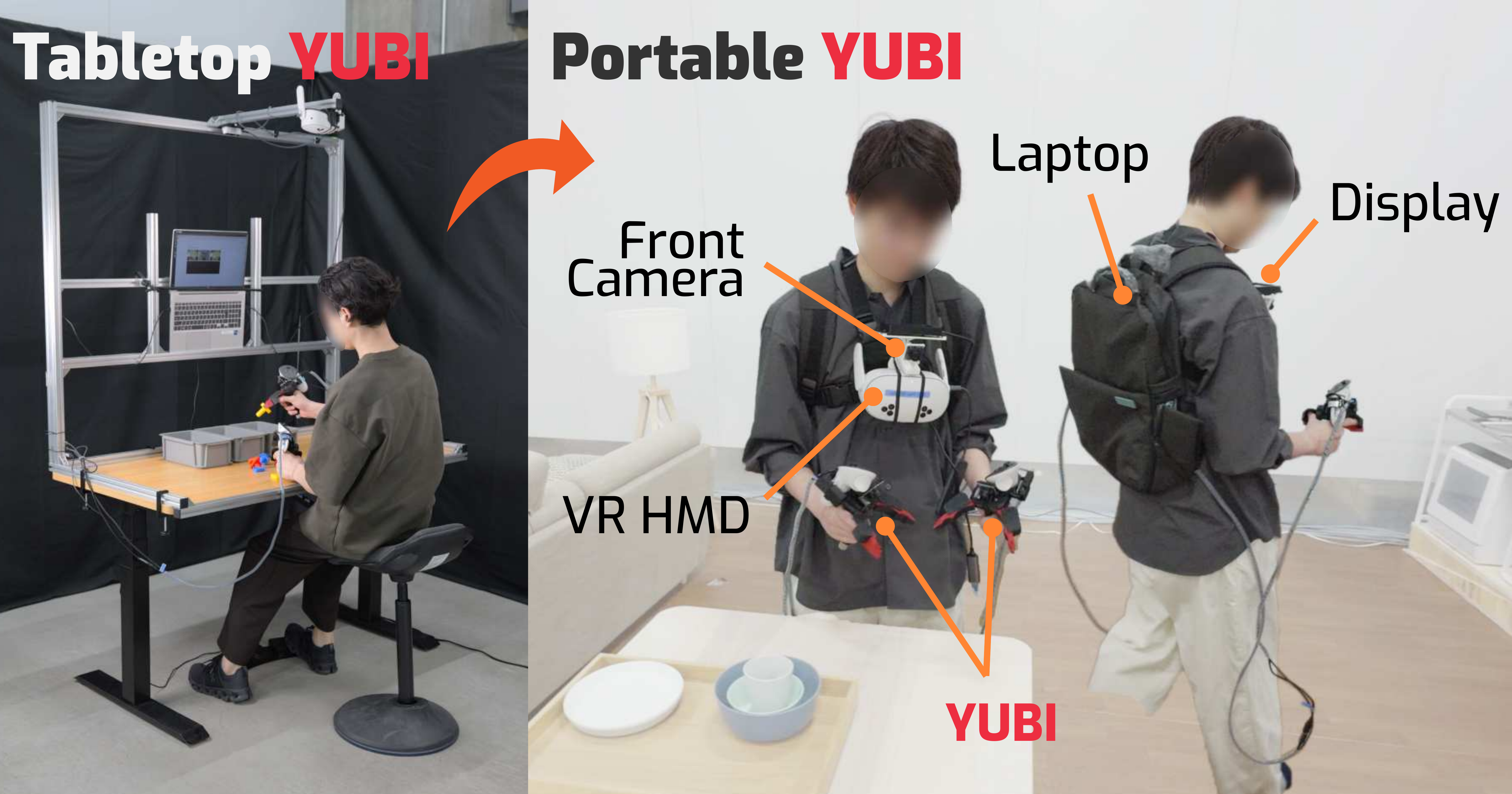}
\caption{\textbf{Portable YUBI configuration.} The operator wears the Quest headset for 6\,DoF controller tracking on the chest, with an egocentric camera replacing the rig-mounted top-view stereo.     
    }
    \label{fig:portable}
\end{figure}

\Cref{fig:portable} illustrates the portable configuration to ship the YUBI data collection to other environments.
In portable mode, the operator detaches the YUBI grippers from the desk rig and wears the Quest 3S headset on the chest-mounted belt.
The other components are adapted for mobility as follows.
The fixed top-view stereo camera is replaced by an egocentric camera mounted on the headset, which captures the workspace from the operator's viewpoint as they move.
The foot pedal cannot be used while walking, so sub-action transitions are instead triggered by double-clicking the YUBI gripper.
The laptop hub is carried in a shoulder bag, while the wrist cameras and gripper hardware remain wired from the hub.
The data schema (\Cref{sec:data_proc}) is preserved across modes, so portable and tabletop episodes share the same downstream pipeline.

\customparagraph{Captured scenarios}
The portable mode unlocks a class of demonstrations that the tabletop rig simply cannot host.
These tasks span more than the workspace volume above a desk.
Illustrative examples we have recorded include
\emph{stacking a cup onto a plate},
\emph{putting away the cooking utensils},
\emph{putting the tray in the dishwasher and closing it},
\emph{putting a book away on the bookshelf},
\emph{putting the remote back on the table},
\emph{hanging a shirt on a hanger}, and
\emph{folding a shirt}.
These are routine actions in real kitchens, dining rooms, and bedrooms; capturing them in their actual settings ensures that the visual context, object positions, lighting, and reach requirements all reflect the conditions a deployed robot would encounter.

\begin{figure}[t]
    \centering
    \begin{minipage}{0.49\linewidth}
        \centering
        \includegraphics[width=\linewidth]{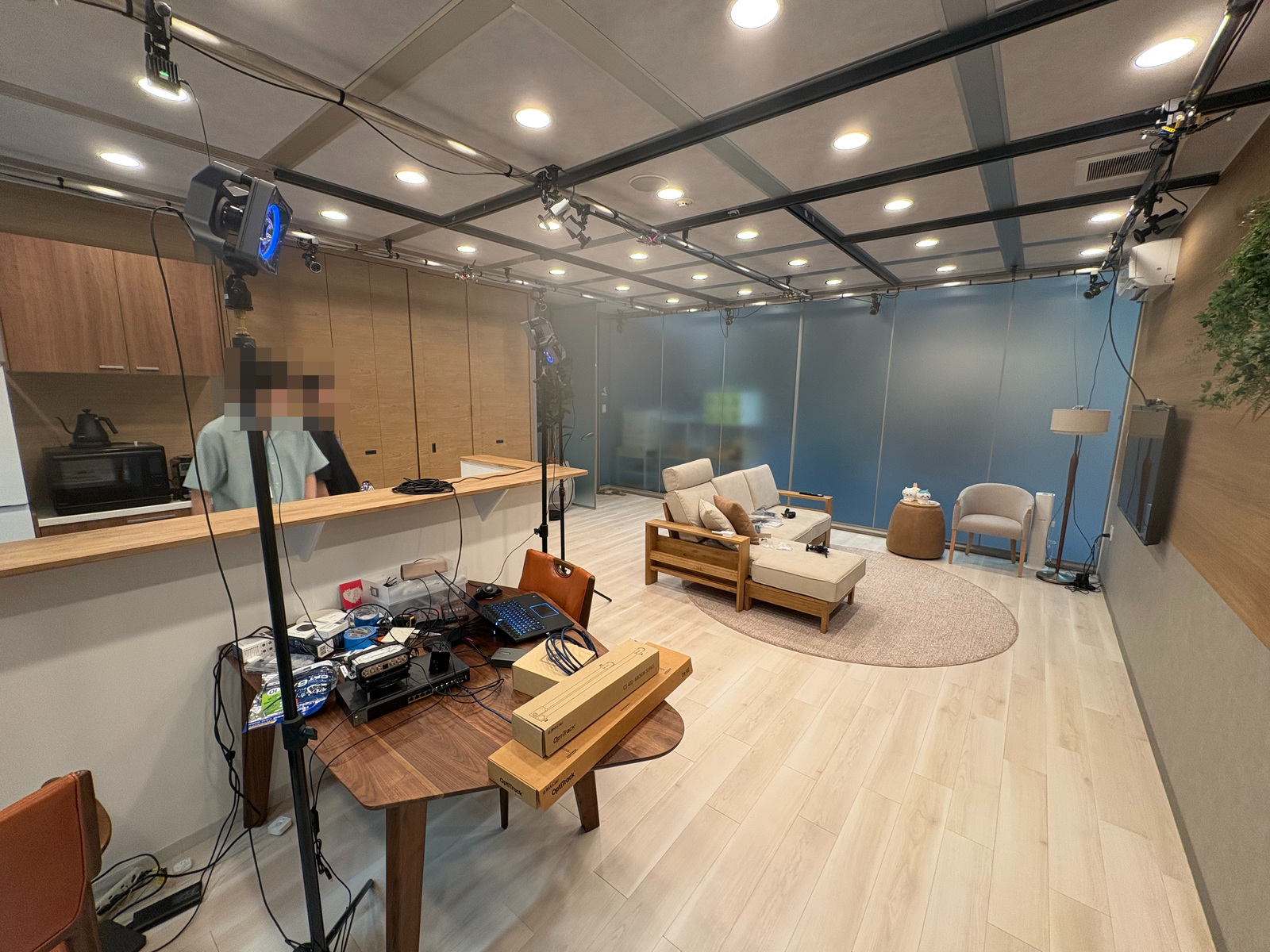}\\
        \small (a) Capture room
    \end{minipage}\hfill
    \begin{minipage}{0.49\linewidth}
        \centering
        \includegraphics[width=\linewidth]{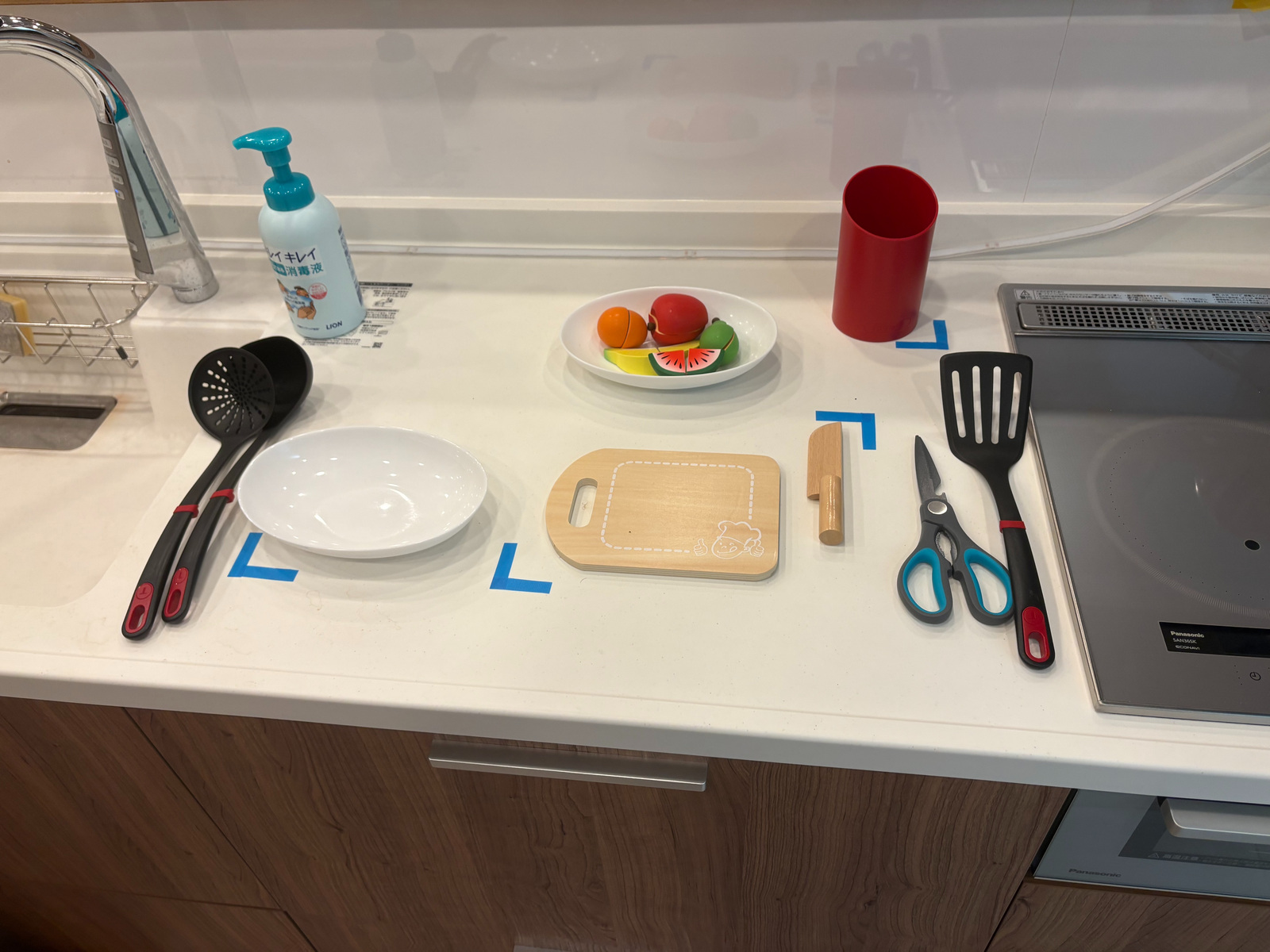}\\
        \small (b) Kitchen task space
    \end{minipage}
    \caption{\textbf{Motion-capture session at AIST~\cite{deguchi2026mapmonoego}.} (a) The household environment, with OptiTrack cameras on ceiling rails and stands. (b) Objects on the kitchen counter, with blue tape marking their initial positions. Each YUBI gripper carries a four-marker rigid body that serves as ground truth for \Cref{supp:tracking_accuracy}. Faces are anonymized.}
    \label{fig:aist_mocap}
\end{figure}

\customparagraph{Motion-capture session}
To evaluate tracking accuracy outside the tabletop rig, we recorded portable YUBI in a household environment at the AIST Digital Human Research Center (AIST Tokyo Waterfront) instrumented with an OptiTrack motion-capture system (\Cref{fig:aist_mocap}).
Each YUBI gripper carries a rigid body of four reflective markers, tracked at 60\,Hz.
In each take, the operator first shows a QR-code clip on a tablet to the wrist cameras and a mocap reference camera for time synchronization, and then performs five household tasks in sequence:
\emph{putting cooking utensils (\eg, a ladle) into a container},
\emph{wiping a cutting board in the sink}, which moves the whole body sideways,
\emph{opening the cap of a plastic bottle},
\emph{placing fruit on a plate}, with the plate set apart to require large arm motions, and
\emph{erasing a whiteboard}.
We recorded three takes of about 2.5\,min each (Trial~1--3), all by one operator, with the Quest~3S.
The accuracy results are reported in \Cref{supp:tracking_accuracy}.

\customparagraph{Outlook}
We see the operator's whole-body 3D trajectory recorded by the chest-mounted Quest as a natural data source for embodiments beyond fixed-base bimanual arms.
Mobile manipulators, wheeled service robots, and humanoid platforms all require coordinated body and end-effector motion in human-scale environments, and the portable YUBI episodes directly provide such paired observations and trajectories.
Investigating further lightweight and robust portable systems with YUBI grippers is a promising avenue we leave for future work.

\section{Experimental Details}\label{sec:deploy_setup}

\subsection{Tracking Accuracy}\label{supp:tracking_accuracy}
We compare Quest controller tracking with UMI-style monocular ORB-SLAM3~\cite{campos2021orbslam3} wrist-camera tracking~\cite{chi2024universal} on the three AIST takes described in \Cref{supp:portableyubi}, using the OptiTrack rigid-body poses as ground truth.

\customparagraph{Protocol}
We use the takes cyclically: each take in turn serves as the calibration take and the other two are evaluated, giving six calibration--evaluation pairs in which every take is evaluated twice.
We fix the hand-eye transform between each mocap rigid body and its Quest controller by pose fitting on the calibration take.
The wrist camera is rigidly mounted on the gripper, so we treat its pose relative to the rigid body as a one-off device calibration: one transform per wrist, estimated from reliable views of a ChArUco board pooled over all three takes, with per-unit intrinsics.
Each take shows the board in only a few short segments, which do not constrain the transform reliably on their own, so the evaluation take's board views also contribute; the calibration data therefore differ between the methods, and only the camera calibration includes the evaluation take.
Leaving the evaluation take out of the camera transform at scoring time increases the SLAM--Quest position difference by 0.2--0.3\,mm; the maps are built with the pooled transform, so this is not a fully out-of-sample calibration.

The wrist camera records at about 99\,Hz; we feed ORB-SLAM3 every third frame (about 33\,fps) at its own timestamp, because ORB-SLAM3 drifts on the full-rate sequence, which we leave to future work.
As in UMI, ORB-SLAM3 loads a prior map of the scene, relocalizes each evaluation take against it and then continues full SLAM with map updates, and the gripper is masked.

Our setup differs from UMI's in four ways.
First, the YUBI wrist camera has no IMU, so ORB-SLAM3 runs monocular rather than monocular-inertial.
Second, as a consequence, the map cannot obtain metric scale from inertial sensing as UMI's does; we instead build it at known camera poses derived from the calibration take's Quest poses, rather than by SLAM on a dedicated mapping video.
(UMI uses an ArUco tag only to express its map in a table frame; our per-take alignment to the ground truth makes this unnecessary.)
Third, the mask follows the gripper's opening angle rather than being static.
Fourth, each wrist has its own map, whereas UMI localizes every gripper camera in a single shared map.

The primary map configuration was specified before evaluation and uses a separate map for each wrist.
The map is built by triangulating ORB features of the calibration take's wrist images at known camera poses, obtained from the calibration take's Quest poses and the hand-eye calibrations, which gives one metric-scale map per wrist.
We test a single map shared by both wrists separately, below.
The comparison is therefore between monocular SLAM that used the Quest at calibration time and the Quest at evaluation time; it does not test whether SLAM can replace the Quest independently.

We score ORB-SLAM3's post-run optimized trajectory, which includes corrections from later frames; the Quest poses are the recorded device outputs.
We ablate the prior map and the gripper mask, and additionally disable loop closing for the map-based setting.
The mask is built from the calibration take by voting on per-frame SAM3~\cite{carion2025sam3} gripper segmentations for each jaw-encoder angle. At each evaluation frame's angle, we discard ORB features inside the mask and along its border rather than blacking out pixels.
Each configuration is run three times per pair and wrist, and all runs are scored.

\customparagraph{Metrics}
For the main comparison we keep, per run, the frames for which both ORB-SLAM3 and Quest output a pose, and align each method to the ground truth per evaluation take by a position-only SE(3) Umeyama fit on those frames.
The map-based runs inherit metric scale from their map, so SE(3) suffices; map-free runs have no metric scale, so we fit Sim(3) separately to each map segment, which favours them.
We report the mean absolute trajectory error (ATE) in position and in rotation, the latter as the angle of $\mathbf{R}_\mathrm{gt}^\top \mathbf{R}_\mathrm{est}$ under the same alignment, on the same frames and at the same times, with the hand-eye transforms held fixed.

For temporal relative pose error (RPE), each scored row $i$ is paired with the scored row $j$ nearest in mocap time to $t_i+\Delta$, within one input interval, for $\Delta=0.1$\,s and $1$\,s; pairs spanning a SLAM map segment boundary, a recording interruption or a ground-truth-invalid frame are excluded.
We report the mean translation norm and rotation angle of $(\mathbf{G}_i^{-1}\mathbf{G}_j)^{-1}(\mathbf{P}_i^{-1}\mathbf{P}_j)$, where $\mathbf{G}$ and $\mathbf{P}$ are ground-truth and estimated poses, without alignment and, like ATE, only on frames where SLAM outputs a pose.

For ATE, RPE and coverage, we average the three runs of a pair with equal weight and then the pairs with equal weight.
The conditional 90\,\% confidence intervals for these metrics use a 2\,s block bootstrap of the evaluation takes, with the same blocks shared by all runs, configurations, pairs and Quest, combined with resampling of the three runs ($B=2000$).
These intervals are conditional on the three takes, the calibration maps, both methods' hand-eye transforms and the time synchronization; it does not include uncertainty in map construction, hand-eye calibration or synchronization, or generalization to other recordings.

For availability, the denominator comprises all input frames with valid ground truth, independent of the method, and frames without a pose count as failures: coverage is the fraction with a pose, and success is the fraction with position error at most 10\,mm after aligning each method on its own available frames.

\customparagraph{Time synchronization}
Wrist-camera frame times are mapped to mocap time using the QR-code clip observed by the wrist cameras and the mocap reference camera (the clip displays a running code, so the cameras need not view it simultaneously).
For the primary comparison, we use this mapping without an additional SLAM offset fitted to the evaluation ground-truth poses.
The QR mapping is independent of the pose-error objective but is not assumed exact.
It is extrapolated beyond the QR observations (estimated standard deviation 2.1--5.9\,ms).
The QR clip is part of the evaluation take's recording, so the synchronization is not out-of-sample.

Quest's device clock has no corresponding QR channel and is mapped to mocap time separately for each take using angular-speed cross-correlation followed by robust pose fitting on that take's mocap data.
The synchronization procedures therefore remain asymmetric.

As a sensitivity analysis, we fit one residual SLAM offset per evaluation take, shared across both wrists, calibration maps and runs, by minimizing the mean position ATE.
Reading SLAM at $t+\tau$, the fitted offsets are $-8.5$, $-8.7$ and $-7.7$\,ms for Trial~1--3.
Their physical cause is unresolved; an approximate rolling-shutter model predicts a contribution of about $-3$\,ms.
Applying these offsets, to position and rotation alike, reduces the aggregate SLAM--Quest position-ATE differences by 0.10\,mm (left) and 0.27\,mm (right) without changing the conclusions based on their conditional 90\,\% confidence intervals.

An earlier, pre-specified diagnostic required the median of the per-run fitted offsets across the three runs of each cell (calibration--evaluation pair and wrist) to have magnitude below 10\,ms, with consistent signs across runs; four of the twelve cyclic main cells exceeded the magnitude threshold. These ATE-fitted offsets do not directly measure QR-clock error.
On 13 October 2026, we changed the reporting policy to retain these results and report the diagnostic failure as a finding rather than use it as a pass/fail criterion.
After inspecting both timing variants, we changed the primary comparison from fitted residual offsets to QR timing and retained the fitted-offset results as sensitivity analyses.

\begin{table*}[!t]
  \centering
\caption{Tracking against OptiTrack ground truth at AIST: mean ATE and 1\,s RPE [conditional 90\,\% CI] on paired frames, with per-take SE(3) alignment for ATE and no alignment for RPE; coverage over all input frames with valid ground truth. Six cyclic pairs, QR timing, three runs per pair. Method differences and their CIs are reported in the text. Arrows indicate the better direction; bold marks the better method per wrist.}
  \label{tab:tracking}
\small
  \setlength{\tabcolsep}{3pt}
  \resizebox{\linewidth}{!}{\begin{tabular}{@{}llccccc@{}}
    \toprule
    \multirow{2}{*}{Wrist} & \multirow{2}{*}{Method} & \multicolumn{2}{c}{ATE} & \multicolumn{2}{c}{RPE ($\Delta t=1$\,s)} & \multirow{2}{*}{Cov. [\%] $\uparrow$} \\
    \cmidrule(lr){3-4}\cmidrule(lr){5-6}
    & & Pos. [mm] $\downarrow$ & Rot. [$^\circ$] $\downarrow$ & Pos. [mm] $\downarrow$ & Rot. [$^\circ$] $\downarrow$ & \\
    \midrule
    \multirow{2}{*}{Left} & ORB-SLAM3 & 15.8 [10.7, 21.6] & 2.15 [1.69, 2.79] & 9.1 [7.4, 11.3] & 1.43 [1.25, 1.64] & 84.9 \\
     & Quest & \textbf{5.3} [5.0, 5.6] & \textbf{1.18} [1.07, 1.30] & \textbf{3.9} [3.5, 4.4] & \textbf{1.07} [0.96, 1.19] & \textbf{99.9} \\
    \midrule
    \multirow{2}{*}{Right} & ORB-SLAM3 & 10.5 [9.1, 11.8] & 1.29 [1.16, 1.48] & 8.2 [7.6, 9.0] & 1.19 [1.11, 1.27] & 86.9 \\
     & Quest & \textbf{5.6} [5.1, 6.3] & \textbf{1.07} [1.03, 1.14] & \textbf{4.7} [4.2, 5.3] & \textbf{0.94} [0.89, 0.99] & \textbf{99.3} \\
    \bottomrule
\end{tabular}}
\end{table*}

\customparagraph{Results}
As shown in \Cref{tab:tracking}, averaging the three runs within each pair and then the six pairs with equal weights, ORB-SLAM3's position ATE exceeds Quest's by 10.5 [5.5, 16.3]\,mm (left) and 4.8 [3.4, 6.3]\,mm (right) on the same frames; the rotation ATE differences are $+0.98$ [$0.50$, $1.58$]$^\circ$ (left) and $+0.22$ [$0.06$, $0.41$]$^\circ$ (right), with conditional 90\,\% CIs in brackets.

The difference depends strongly on the calibration map.
With the maps built from Trial~1, SLAM's per-pair mean position ATE on the same frames is 6.6--7.6\,mm, against 4.8--6.9\,mm for Quest, whereas the three weakest pair--wrist combinations reach 20.7--27.0\,mm, against 4.2--6.2\,mm (left wrist, Trial~2 to Trial~3 and to Trial~1; right wrist, Trial~3 to Trial~2; point estimates).
On the left wrist, 4 of the 18 runs have mean position ATE above 30\,mm and raise the aggregate mean; on the right wrist, all three runs calibrated on Trial~3 and evaluated on Trial~2 have SLAM mean position ATE of 17--25\,mm.
We have not identified the cause.

Replacing Quest's scoring transform by a relative-motion hand-eye calibration (pooled over the three takes, without the evaluation take, or from the calibration take alone), with the maps and SLAM runs unchanged, moves the two position differences by at most 0.2\,mm.
The rotation differences are more sensitive to this choice: the right-wrist difference drops to $+0.09$--$0.11^\circ$, with intervals that contain zero, so we draw no conclusion on rotation for the right wrist; the left-wrist difference ($+0.88$ to $+0.98^\circ$) remains positive with confidence intervals excluding zero under every choice.
Removing a constant rotation offset per run, as a further sensitivity check, gives rotation differences of $+0.87^\circ$ (left) and $+0.13^\circ$ (right).

Over short horizons, the gap persists without alignment and is less dependent on the calibration map than ATE: at $\Delta=0.1$\,s, translation RPE is 3.4--3.8\,mm for SLAM versus 0.8--1.0\,mm for Quest, and at $\Delta=1$\,s it is 8.2--9.1\,mm versus 3.9--4.7\,mm.
For $\Delta=1$\,s, the SLAM $-$ Quest RPE differences (left / right; conditional 90\,\% CI) are $+5.2$ [$3.6$, $7.3$] / $+3.5$ [$2.8$, $4.3$]\,mm in translation and $+0.36$ [$0.23$, $0.53$] / $+0.25$ [$0.18$, $0.32$]$^\circ$ in rotation.
At $\Delta=0.1$\,s, the SLAM $-$ Quest RPE differences (left / right; conditional 90\,\% CI) are $+2.6$ [$2.3$, $3.0$] / $+2.8$ [$2.6$, $3.0$]\,mm in translation and $+0.12$ [$0.09$, $0.16$] / $+0.09$ [$0.06$, $0.11$]$^\circ$ in rotation.
All eight temporal RPE differences have conditional 90\,\% confidence intervals excluding zero.

The left-to-right relative pose error is 29.5\,mm / $2.75^\circ$ for SLAM versus 11.4\,mm / $1.72^\circ$ for Quest (difference: $+18.2$ [$14.0$, $22.6$]\,mm / $+1.03$ [$0.55$, $1.61$]$^\circ$), including SLAM's registration error between per-wrist maps (about 14\,mm on Trial~1 maps and 40--45\,mm on the weakest pairs), with the wrists matched within half an input interval rather than strictly simultaneously and Quest read at the same times.
Availability separates the two methods more clearly: ORB-SLAM3 provides a pose for 85--87\,\% of the frames (67\,\% in the worst run), compared with 99--100\,\% for Quest.

The recordings contain one to three interruptions of approximately 2.1--2.6\,s per take, plus a 0.344\,s gap in Trial~3; they affect every recorded stream, including Quest.
Intervals without wrist-camera images are excluded from both methods' availability denominators, and gap lengths are missing-frame counts multiplied by the median input interval, excluding recording interruptions and periods without valid ground truth.
The longest missing-pose stretch per run corresponds to 8--11\,s on average and up to 18\,s by this measure.
Some SLAM pose-loss episodes, including the longest reported stretch, begin immediately after image recording resumes; this temporal association does not establish that the interruption caused the loss.

ORB-SLAM3's success rate within 10\,mm is 51\,\% (left) and 59\,\% (right), compared with 92\,\% and 87\,\% for Quest.
The losses concentrate in placing fruit on the left wrist and erasing the whiteboard on the right wrist (\Cref{tab:tracking_pa}).

In the ablation (\Cref{tab:tracking_ablation}), each ablated and main run with the same run index is scored on their shared pose frames and aligned separately; removing the mask with loop closing on changes position and rotation ATE by $+13.8$ [$2.2$, $27.9$]\,mm and $+0.84$ [$0.05$, $1.99$]$^\circ$ on the left, and $+9.9$ [$2.8$, $17.4$]\,mm and $+1.83$ [$0.60$, $3.31$]$^\circ$ on the right.
Removing the prior map with the mask and loop closing on changes position and rotation ATE by $+25.5$ [$9.9$, $36.7$]\,mm and $+3.72$ [$2.87$, $5.78$]$^\circ$ on the left, and $+35.2$ [$19.0$, $48.2$]\,mm and $+8.55$ [$5.26$, $12.90$]$^\circ$ on the right; all these paired conditional 90\,\% CIs exclude zero.

Disabling loop closing with the mask reduces position error by 2.7\,mm on the right (paired CI: [$-4.1$, $-1.4$]\,mm), but its interval contains zero on the left ($-4.9$ [$-11.5$, $1.7$]\,mm); both rotation intervals contain zero (left $-0.44$ [$-1.00$, $0.02$]$^\circ$, right $-0.12$ [$-0.31$, $0.02$]$^\circ$).
On the two pairs mapped from Trial~1 it gives no gain (left $+5.1$ [$-2.1$, $15.8$]\,mm, right $+0.3$ [$-0.2$, $0.8$]\,mm).
This suggests, but does not establish, a benefit from disabling loop closing, and the main configuration keeps loop closing on, as specified before the evaluation.

\begin{table*}[!t]
  \centering
\caption{ORB-SLAM3 (six cyclic pairs, QR timing, three runs/pair): ATE/RPE means [marginal conditional 90\,\% CI] on own frames; coverage on GT-valid inputs. Low coverage can flatter ATE; compare paired differences in text, never CI overlap. Bold: best map-based SE(3) means, including displayed ties. Map-free rows: favourable Sim(3) per segment, unbolded; 1\,s RPE translation uses that segment's scale.}
  \label{tab:tracking_ablation}
  \small
\setlength{\tabcolsep}{3pt}
  \resizebox{\linewidth}{!}{\begin{tabular}{@{}lcccccccc@{}}
    \toprule
\multirow{2}{*}{Wrist} & \multirow{2}{*}{Map} & \multirow{2}{*}{Mask} & \multirow{2}{*}{Loop} & \multicolumn{2}{c}{ATE} & \multicolumn{2}{c}{RPE ($\Delta t=1$\,s)} & \multirow{2}{*}{Cov. [\%] $\uparrow$} \\
    \cmidrule(lr){5-6}\cmidrule(lr){7-8}
    & & & & Pos. [mm] $\downarrow$ & Rot. [$^\circ$] $\downarrow$ & Pos. [mm] $\downarrow$ & Rot. [$^\circ$] $\downarrow$ & \\
    \midrule
\multirow{6}{*}{Left} & $\checkmark$ & $\checkmark$ & $\checkmark$ & 15.8 [10.7, 21.6] & 2.15 [1.69, 2.79] & 9.1 [7.4, 11.3] & 1.43 [1.25, 1.64] & 84.9 \\
    & $\checkmark$ &  & $\checkmark$ & 31.0 [16.9, 47.2] & 2.91 [2.06, 4.24] & 13.0 [9.5, 17.1] & 1.54 [1.33, 1.79] & 83.6 \\
    &  & $\checkmark$ & $\checkmark$ & 46.9 [33.4, 56.1] & 8.61 [7.26, 11.72] & 20.6 [16.2, 25.0] & 1.96 [1.64, 2.32] & 86.2 \\
    &  &  & $\checkmark$ & 96.8 [67.0, 113.7] & 26.15 [20.30, 35.24] & 41.7 [31.3, 52.0] & 3.11 [2.51, 3.80] & 83.2 \\
    & $\checkmark$ & $\checkmark$ &  & \textbf{10.7} [7.1, 15.3] & \textbf{1.67} [1.45, 2.01] & \textbf{8.4} [6.6, 10.8] & \textbf{1.40} [1.22, 1.60] & 87.4 \\
    & $\checkmark$ &  &  & 59.0 [34.4, 80.6] & 5.60 [3.63, 7.90] & 15.1 [10.9, 20.4] & 1.80 [1.48, 2.18] & \textbf{88.0} \\
    \midrule
    \multirow{6}{*}{Right} & $\checkmark$ & $\checkmark$ & $\checkmark$ & 10.4 [9.1, 11.8] & \textbf{1.29} [1.16, 1.48] & 8.2 [7.6, 9.0] & \textbf{1.18} [1.11, 1.26] & 86.9 \\
    & $\checkmark$ &  & $\checkmark$ & 19.9 [12.8, 27.5] & 3.14 [1.94, 4.64] & 12.1 [10.0, 14.7] & 1.64 [1.39, 1.94] & 76.3 \\
    &  & $\checkmark$ & $\checkmark$ & 48.2 [32.6, 59.8] & 11.34 [7.89, 15.73] & 27.5 [20.6, 34.0] & 1.55 [1.36, 1.79] & 87.3 \\
    &  &  & $\checkmark$ & 77.8 [52.2, 90.8] & 24.27 [15.90, 32.91] & 43.3 [31.6, 53.5] & 2.64 [2.15, 3.19] & 81.3 \\
    & $\checkmark$ & $\checkmark$ &  & \textbf{8.4} [7.4, 10.0] & \textbf{1.29} [1.12, 1.63] & \textbf{8.0} [7.3, 8.7] & 1.23 [1.14, 1.33] & \textbf{93.8} \\
    & $\checkmark$ &  &  & 27.9 [8.7, 51.5] & 3.62 [1.41, 7.08] & 12.6 [8.7, 17.5] & 1.49 [1.31, 1.72] & 92.6 \\
    \bottomrule
\end{tabular}}
\end{table*}

\begin{table*}[!t]
  \centering
\caption{Task pose coverage [\%] of GT-valid inputs (six cyclic pairs; ORB-SLAM3: main, three runs/pair). Higher is better; bold: wrist/task maxima, including ties.}
  \label{tab:tracking_pa}
  \small
  \resizebox{\linewidth}{!}{\begin{tabular}{ll ccccccc}
    \toprule
    Wrist & Method & QR sync & Utensils & Cutting board & Bottle cap & Fruit & Whiteboard & Transitions \\
    \midrule
\multirow{2}{*}{Left} & ORB-SLAM3 & \textbf{100} & 95 & 94 & 67 & 35 & 86 & 76 \\
      & Quest & \textbf{100} & \textbf{100} & \textbf{100} & \textbf{100} & \textbf{100} & \textbf{100} & \textbf{100} \\
    \midrule
\multirow{2}{*}{Right} & ORB-SLAM3 & \textbf{100} & 95 & 85 & 90 & \textbf{100} & 72 & 93 \\
      & Quest & \textbf{100} & \textbf{100} & \textbf{100} & \textbf{100} & 99 & \textbf{99} & \textbf{97} \\
    \bottomrule
  \end{tabular}}
\end{table*}

All 198 runs completed; 16 were re-run after a crash or hang.
UMI uses thresholds of 60 cumulative tracking failures to abort a demonstration and more than 10 lost frames to discard it.
Counting, in our $\sim$33\,fps output, the input frames after the first pose that have no pose, the main configuration loses 79--1339 frames per run (median 575.5), so all 36 of its runs reach both thresholds; over all 198 runs, 197 reach the abort threshold and all reach the discard threshold.
This is an offline proxy using UMI's frame-count thresholds, not a reproduction of its online tracking-state checks. For map-free runs, it uses the exported single-map trajectory.

\customparagraph{Map shared by both wrists}
In an additional exploratory analysis, we evaluated a single map per calibration take built from both wrists' keyframes, as in UMI, using the same scoring rules.
It relocalizes and tracks both wrists.
Relative to the per-wrist maps, it changes the position error by $-2.1$ [$-3.6$, $-0.8$]\,mm on the right wrist and $-0.4$ [$-5.9$, $5.9$]\,mm on the left.
It lowers the error on the per-wrist maps' weakest pairs (right wrist, Trial~3 to Trial~2: 17--25\,mm per run become 9--11\,mm; left wrist, Trial~2 to Trial~3: 16--33\,mm become 7--9\,mm), but one left-wrist run calibrated on Trial~2 and evaluated on Trial~1 has a mean position ATE of 115\,mm, against at most 56\,mm for the per-wrist map on this pair.

On the two pairs mapped from Trial~1, where the per-wrist maps already perform well, it is worse on both wrists (left $+3.2$ [$0.4$, $5.9$]\,mm, right $+1.2$ [$0.2$, $2.2$]\,mm), and the right-wrist coverage drops from 86\,\% to 79\,\%.
The shared map is not a uniform improvement; we retain the prespecified per-wrist maps for the primary comparison and report the shared-map results as an exploratory comparison.

\customparagraph{Quest alone}
Scored on every controller sample at its device time mapped to mocap time, rather than at the wrist-camera frames, Quest's sample-weighted position error over the six pairs has a mean of $5.3$ [$4.9$--$5.6$]\,mm (left) and $6.5$ [$5.4$--$8.3$]\,mm (right) [conditional 90\,\% CI], and a median of $4.6$ and $4.9$\,mm, respectively. Coverage is 99.2--100\,\%, and mean rotation error is $1.16^\circ$ (left) and $1.12^\circ$ (right); changing the calibration take moves a pair's mean position error by at most $0.27$\,mm.

The right-wrist mean is raised by a single event in Trial~2 (placing fruit): for about 3.6\,s the Quest pose deviates by up to 129\,mm from the mocap, which is continuous there, while the controller is still reported as tracked; a Quest-side tracking error is the likely cause.
Excluding these samples from the pair calibrated on Trial~1 and evaluated on Trial~2 would reduce its right-wrist mean position error from $6.6$ to $4.6$\,mm. We retain them in the score.
Such events are what the filtering in \Cref{sec:data_proc} targets, but this one was not flagged by Quest's tracking state.

\subsection{Usability Test Protocol}
\begin{figure}
\centering
\includegraphics[width=\hsize]{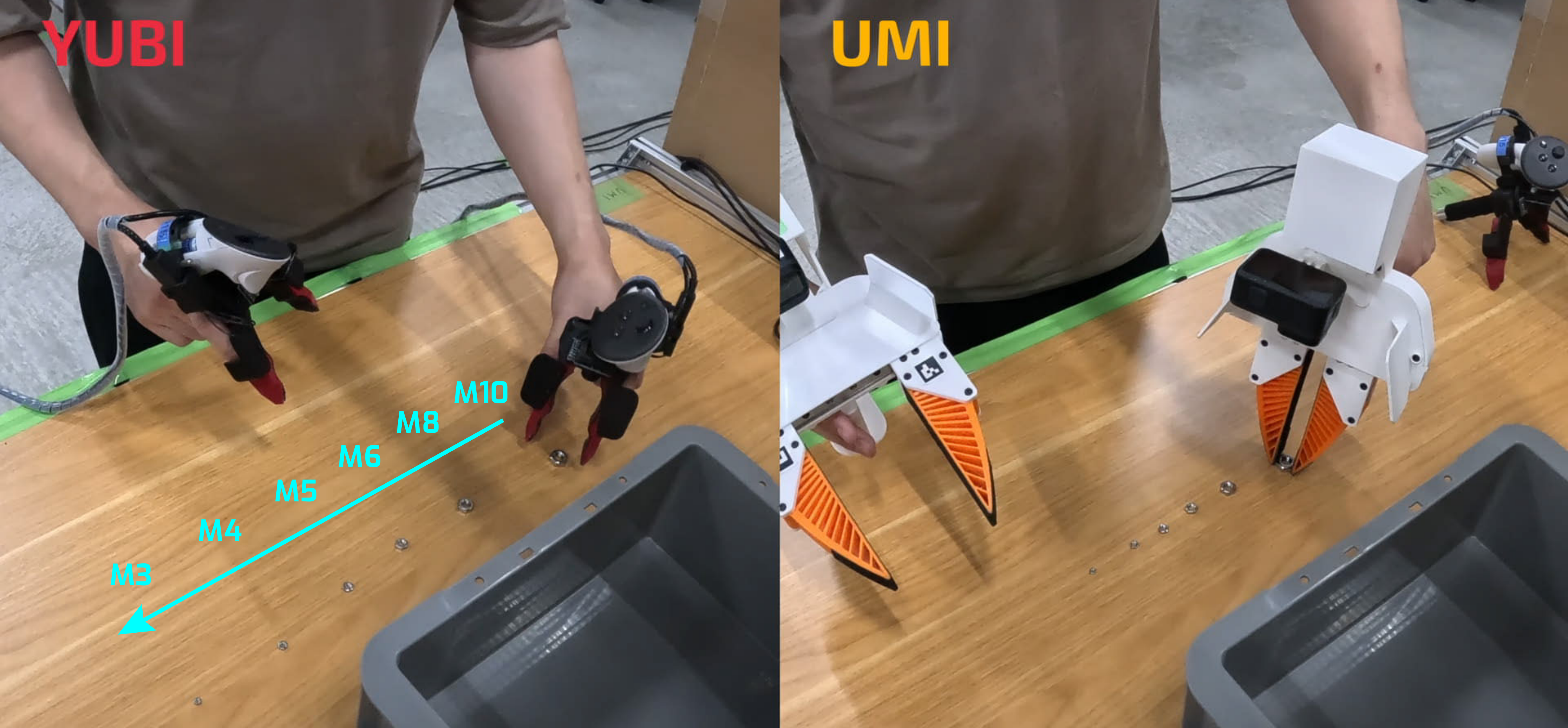}
\vspace{-2mm}
\captionof{figure}{\textbf{Dexterity-test protocol.} Single-attempt pick-and-place of M10--M3 hex nuts. UMI's bulky jaws occlude small parts and crowd the workspace, whereas YUBI's slim fingertips reach the nut with a pinch-like contact.}
\label{fig:dexterity_test_howto}
\end{figure}

\customparagraph{Dexterity test}
\Cref{fig:dexterity_test_howto} illustrates the dexterity-test protocol using either YUBI or UMI.
To quantitatively evaluate this advantage, we conducted a user study assessing dexterity in grasping objects of varying sizes using nuts.
Here, operators were instructed to pick up a nut from a table and place it into a tray.
Six standardized hex nut sizes (M10--M3) were prepared, and operators performed the pick-and-place task sequentially from the largest to the smallest.
Note that operators had only one attempt per nut. If they failed to grasp or place it, it was counted as a failure, and they proceeded to the next size.
A total of 10 operators were recruited with gender balanced, each performing five attempts.

\Cref{fig:dexterity_test} compares the success rates of UMI and YUBI. On the larger nuts (M8 and M10), both devices approach the ceiling ($\ge 94\%$). As the diameter decreases, however, the two methods diverge. YUBI maintains a clear margin over UMI at M6 ($+20$~pp) and M5 ($+10$~pp), and the gap widens most markedly at the smallest size: on M3 nuts, YUBI achieves a $44\%$ success rate while UMI drops to $14\%$, a roughly $3\times$ improvement.
The sole exception to this trend appears at M4, where YUBI's success rate dips below its own M3 rate.
We hypothesize that a geometric mismatch exists between the nut diameter and the fingertip curvature, \ie, a size-specific contact artifact rather than a loss of precision.
Indeed, we find that UMI also exhibits the same dip from M6$\rightarrow$M5, supporting this interpretation.
Overall, these results suggest that YUBI is more adaptable to precision tasks.

\customparagraph{Operational efficiency test}
We further demonstrate that the intuitive design and improved manipulability of YUBI enable faster data collection compared to existing interfaces.
We measured the time required to complete a task for five distinct tasks. Each participant was assigned two tasks to use either manual execution (Hand), UMI~\cite{chi2024universal}, and YUBI. For each task-device combination, operators completed five trials, and the average completion time was computed across trials. A total of 10 operators were recruited, with gender balanced, resulting in four operators assigned to each task. To mitigate order effects, operators were counterbalanced such that half performed the conditions in the order Hand$\to$UMI$\to$YUBI, while the remaining half followed vice versa.

\Cref{fig:efficiency_test} shows the average task completion time of each device on the five evaluation tasks.
Across all tasks, YUBI consistently outperforms UMI, with speed-up ratios ranging from $1.37\times$ on the domino arrangement task to $4.19\times$ on the phone-charging task.
The results demonstrate that YUBI substantially narrows the efficiency gap toward the human hand compared with UMI, with clear advantages on tasks that require fine and dexterous manipulation, such as connector insertion in the phone-charging task and precise tool control in the whiteboard-writing task.

\subsection{Additional Physical Studies}\label{supp:physical_studies}

\begin{figure}[t]
\centering
\includegraphics[width=\hsize]{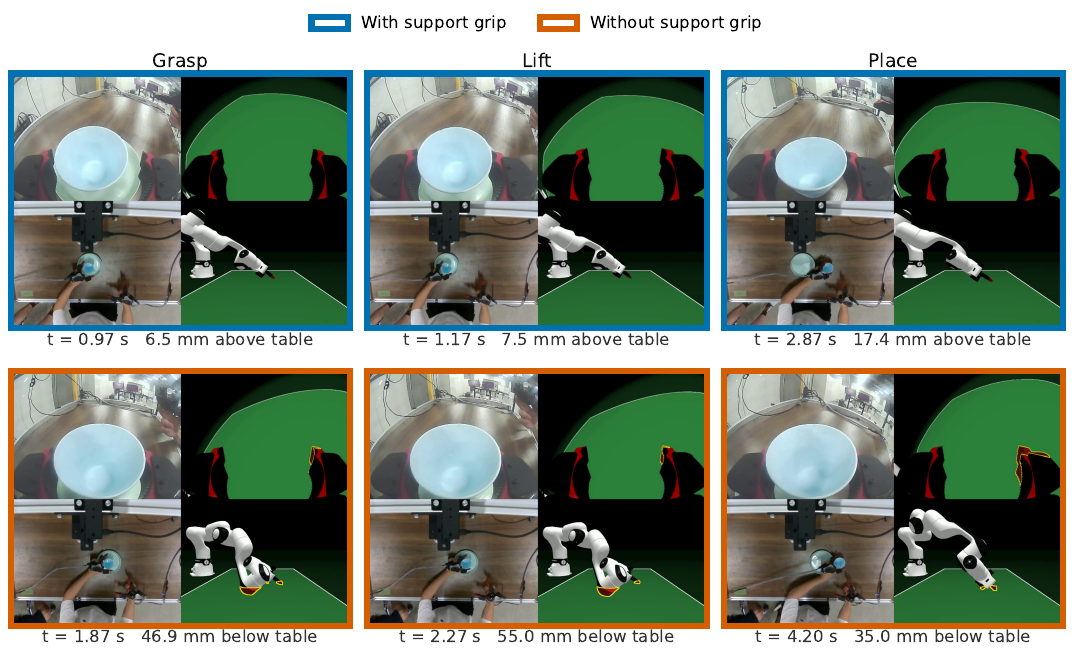}
\vspace{-5mm}
\caption{Kinematic replay of the same primitive action with (blue) and without (orange) the support grip.
We selected the least-penetrating episode of each condition and, within the grasp, lift and place windows, the frame in which the arm is lowest.
Red areas with yellow outlines indicate penetration below the table surface ($z=0$).}
\label{fig:support_grip_min_pair}
\end{figure}

\begin{table}[t]
\centering
\caption{Effect of the support grip on the replayed arm (Cup task, 7 paired demonstrations).
Values are mean $\pm$ SD (median). $p$: paired sign-flip permutation test.}
\label{tab:support_grip_summary}
\small
\setlength{\tabcolsep}{5pt}
\begin{tabular}{lccc}
\toprule
Metric & With support grip & Without support grip & $p$ \\
\midrule
TCP min [mm]               & $53.0 \pm 9.0$ (53.9)   & $38.3 \pm 5.1$ (37.5)   & 0.016 \\
Max penetration [mm]       & $13.7 \pm 12.2$ (12.6)  & $45.7 \pm 17.2$ (55.0)  & 0.016 \\
Penetration frames [\%]    & $22.4 \pm 14.4$ (19.5)  & $48.5 \pm 10.5$ (51.4)  & 0.031 \\
Episodes with penetration  & 6/7                     & 7/7                     & --    \\
\bottomrule
\end{tabular}
\end{table}

\customparagraph{Support grip and motion transfer}
To evaluate the effect of the support grip, we recorded seven paired cup-task demonstrations with and without the grip, retargeted both through the same bimanual Franka IK pipeline, and replayed them kinematically against the table plane ($z=0$).
Without the support grip, the minimum TCP height decreased from 53.0 to 38.3\,mm, the maximum penetration into the table increased from 13.7 to 45.7\,mm, and the fraction of penetrating frames increased from 22.4 to 48.5\,\% (Table~\ref{tab:support_grip_summary}; paired sign-flip permutation test, $p \le 0.031$).
Fig.~\ref{fig:support_grip_min_pair} shows that, even in the best case, the arm without the support grip sinks into the table during grasp, lift, and place.
These results support using the support grip for motion transfer.

\begin{figure}[t]
    \centering
    \begin{minipage}[t]{0.70\linewidth}
        \centering
        \includegraphics[width=\linewidth]{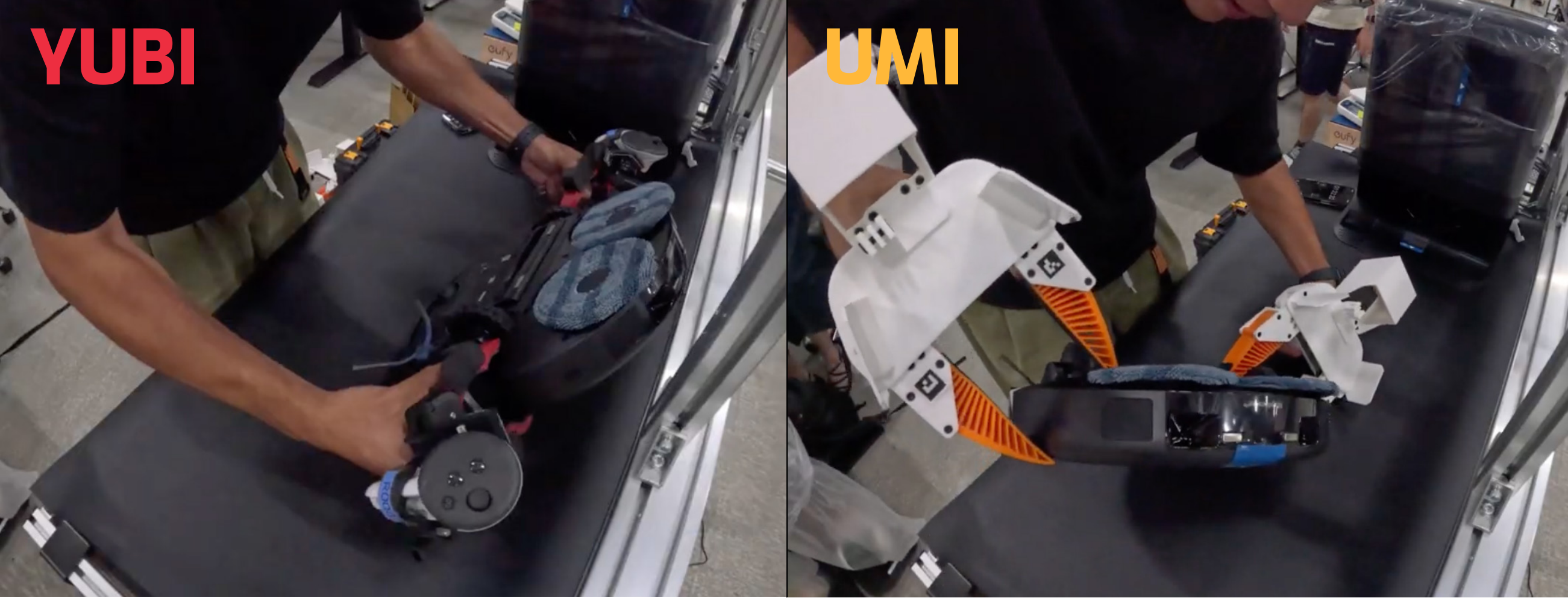}
    \end{minipage}
    \hfill
    \begin{minipage}[t]{0.29\linewidth}
        \centering
        \includegraphics[width=\linewidth]{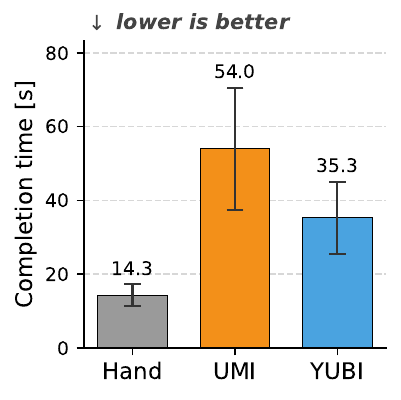}
    \end{minipage}
    \caption{\textbf{Heavy-object handling.} \textbf{Left:} Participants are required to manipulate a 3.3\,kg vacuum using YUBI and UMI. \textbf{Right:} Task completion time with Hand/UMI/YUBI.}
    \label{fig:heavy_object}
\end{figure}

\customparagraph{Heavy-object handling}
In an additional robot-vacuum maintenance study under the same Hand/UMI/YUBI conditions, operators handled a 3.3\,kg vacuum (Fig.~\ref{fig:heavy_object}, left). YUBI sustained the load in all 25 trials without a drop. Mean completion time was 35.3\,s with YUBI, 54.0\,s with UMI, and 14.3\,s with bare hands, corresponding to a 34.6\% reduction relative to UMI (Fig.~\ref{fig:heavy_object}, right). This provides task-specific evidence for heavy-object handling efficiency.

\begin{figure}[t]
    \centering
    \begin{minipage}[t]{0.48\linewidth}
        \centering
        \includegraphics[width=1\linewidth]{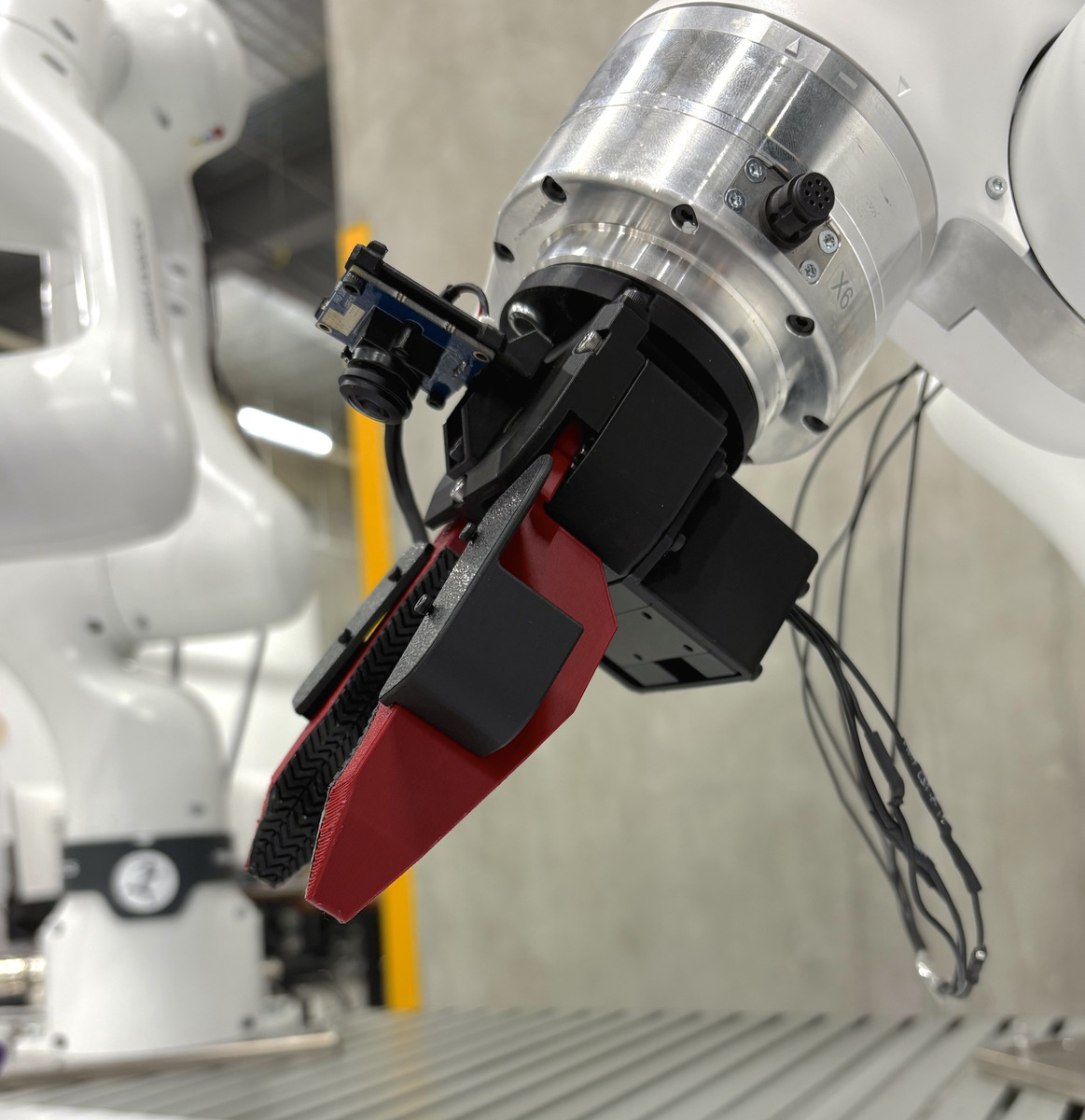}
    \end{minipage}\hfill
    \begin{minipage}[t]{0.48\linewidth}
        \centering
        \includegraphics[width=1\linewidth]{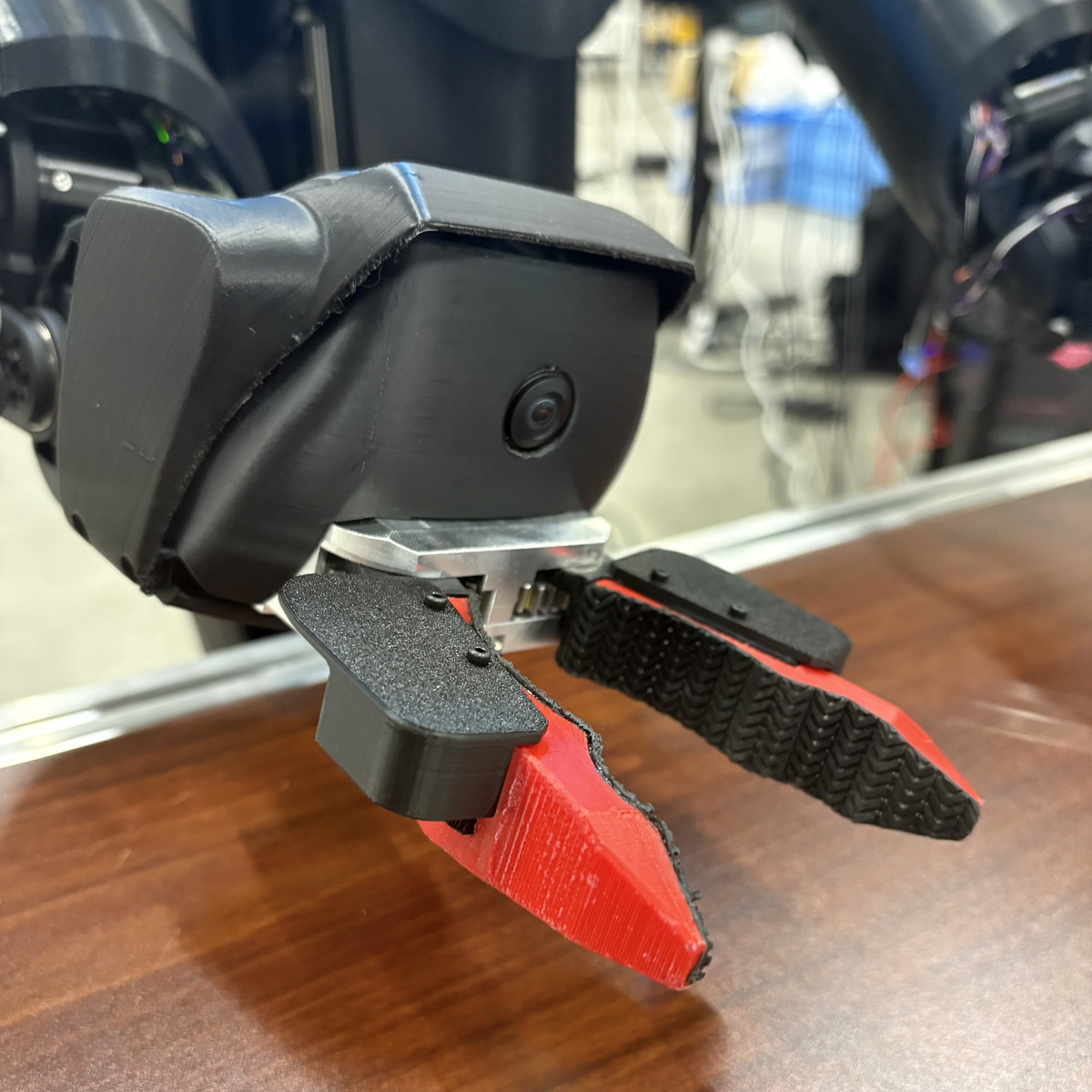}
    \end{minipage}
    \caption{\textbf{YUBI gripper mounted on the deployment robots (Left: Franka, Right: ELEY).} The robot's native gripper is removed and replaced with the YUBI gripper through a per-robot adapter flange; a motor drives the jaw aperture via the same actuation gear used during data collection.}
    \label{fig:yubi_mount}
\end{figure}

\subsection{Robot Platforms}
We deploy YUBI-trained policies on three bimanual robot platforms with distinct kinematics: UR~\cite{ur2026}, Franka~\cite{franka2026},
and Toyota semi-humanoid ELEY~\cite{eley2026}.
On each platform, we physically replace the robot's native gripper with the YUBI gripper and attach the same wrist camera, so that the deployed end-effector matches the one used during data collection. Since the YUBI gripper is passively finger-driven during data collection, we attach a motor to its actuation gear for deployment, so that the jaw aperture can be driven by an open/close command. The recorded per-hand apertures $d^{\text{right}}, d^{\text{left}}$ are then linearly mapped to this motor command.

\Cref{fig:yubi_mount} shows the resulting end-effector on two of the target platforms.
We mount YUBI so that the robot's wrist-flange rotation axis passes straight through the gripper's pinch point, aligned with the fingers' approach direction.
With this alignment, a pure wrist roll spins the fingertips in place rather than swinging them around an off-center pivot, and because the fingertips sit on the wrist axis with no lateral offset, the remaining joints are not consumed compensating for that offset.
The arm can therefore use its full kinematic range to follow the recorded end-effector trajectory.

\subsection{Policy Training and Inference}
We fine-tune $\pi_{0.5}$ on the YUBI demonstrations and evaluate the resulting policy on each deployment task.
The post-training data comprise $408$, $275$, $194$, $1{,}903$, $1{,}691$, $3{,}985$ demonstrations for ``ball in basket'', ``stack cup pyramid'', ``unfold glasses'', ``pick and place socks'', ``tape in box'', ``cup placement'', respectively.
We use a task-balanced data sampler to avoid overfitting or underfitting to specific tasks.
For the experiments using UR, we train 30K iterations with batch size of 256 and action chunk of 16.
For the experiment using Franka and ELEY, we train 150K iterations with batch size of 64 and action chunk of 32.

\subsection{Comparison to Diffusion Policy}
\label{sec:dp_compare}
We ablate the $\pi_{0.5}$~\cite{black2025pi05} policy against the diffusion policy (DP) architecture~\cite{chi2023diffusionpolicy} originally adopted by UMI~\cite{chi2024universal}: a frozen CLIP image and text encoder feeding a diffusion-based action decoder trained from scratch on the YUBI demonstrations to predict relative end-effector action chunks. To keep the comparison fair, both models are trained per task on the corresponding YUBI demonstrations and evaluated under the same protocol on the bimanual UR platform (\Cref{table:dp_compare}). On the simplest task, \emph{ball in basket}, both saturate at $20/20$. The gap opens on \emph{stack cup pyramid} ($9/20$ vs $13/20$), where the second placement must be executed under stacking-induced contact, and widens to a categorical failure on \emph{unfold glasses} ($0/20$ vs $9/20$), the most asymmetric bimanual task in the suite.

\begin{table}[t]
  \caption{\textbf{Model ablation on the bimanual UR platform.} Success rate (out of 20 rollouts per task) for the diffusion-policy baseline~\cite{chi2023diffusionpolicy} and our $\pi_{0.5}$~\cite{black2025pi05} policy, each fine-tuned per task on the corresponding YUBI demonstrations.}
  \vspace{-2mm}
  \label{table:dp_compare}
  \centering
  \small
  \begin{tabular*}{\linewidth}{@{\extracolsep{\fill}}lccc@{}}
    \toprule
    \textbf{Model} & \textbf{Ball in basket} & \textbf{Stack cup pyramid} & \textbf{Unfold glasses} \\
    \midrule
    DP~\cite{chi2023diffusionpolicy} (CLIP Enc.)  & 20/20 & 9/20  & 0/20 \\
    $\pi_{0.5}$~\cite{black2025pi05}                              & 20/20 & 13/20 & 9/20 \\
    \bottomrule
  \end{tabular*}
\end{table}

This pattern motivates our choice of $\pi_{0.5}$ as the primary policy. With the same per-task data, a from-scratch diffusion decoder is sufficient for the easiest task but lacks the prior knowledge required to recover as the task becomes harder. In contrast, $\pi_{0.5}$ inherits a vision-language-action backbone pretrained on large-scale robot data, which provides a strong inductive bias for visual representations and contact-aware end-effector trajectories. We therefore adopt $\pi_{0.5}$ throughout this paper and treat DP as a structural lower bound.

\section{Open-Source Release}\label{sec:open_source}

To support reproduction and community extension, we release the complete YUBI stack, including hardware, data-collection software, and the curated dataset together with the paper. 
All operators consented to data collection, and their faces are anonymized in the released data.

\subsection{Hardware}

\begin{figure}
\centering
\includegraphics[width=0.5\hsize]{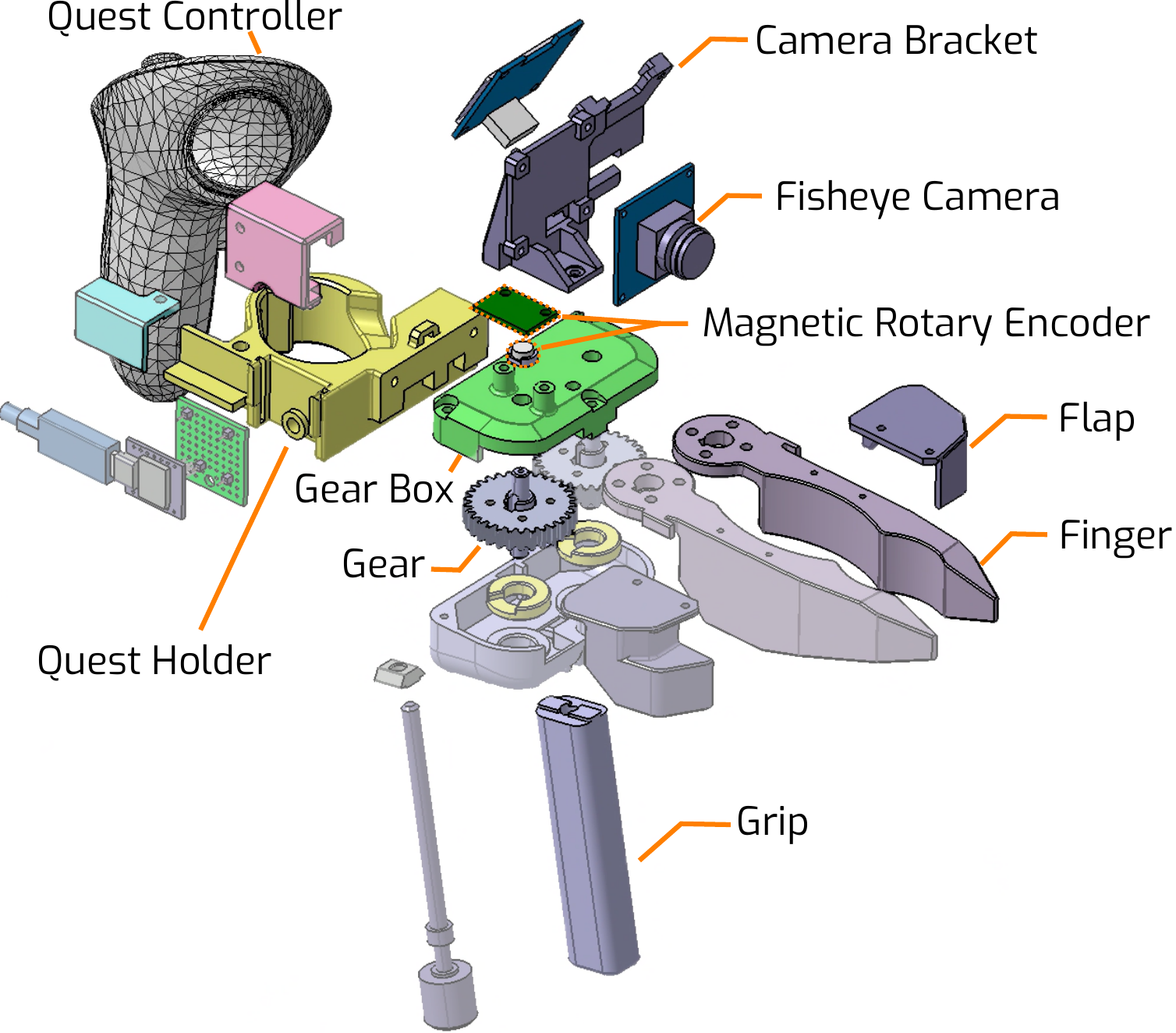}
\captionof{figure}{
\textbf{Exploded view of the YUBI gripper.}
The bidigital mechanism utilizes internal gears to actuate the jaws, all supported by an ergonomic grip and flap. A fisheye camera is attached for task observation. The Quest controller is integrated to provide high-frequency 6-DoF trajectory tracking.
}
\label{fig:yubi}
\end{figure}

The hardware release covers every part needed to fabricate the YUBI system end-to-end: STEP CAD assemblies for the YUBI gripper (see \Cref{fig:yubi}), the robot-mounted gripper (with the DYNAMIXEL-actuated jaw used for deployment), the stationary data-collection desk rig, and the portable collection rig; STL files for every 3D-printable component; a bill of materials in CSV format with sourcing notes; PDF assembly guides of the gripper, and rig builds; and adapter flanges that interface the gripper with each robot platform used in our deployment experiments (\Cref{sec:deploy_setup}). The release is distributed under the CERN Open Hardware Licence~v2 (Weakly Reciprocal), permitting both reproduction and downstream modification.

\subsection{Software}
The software release contains the operator-side bringup stack used to capture the \DBhours{} hours of demonstrations reported in this paper, distributed under the Apache License~2.0.
It is implemented in Python with ROS\,2 nodes and ESP32-C6 firmware, and packaged as a multi-container Docker stack.
A web application for data collection management, with frontend and backend written in Go and TypeScript, is also included.
The release ships drivers and launch configurations for every signal source described in \Cref{sec:op_setup}.
These cover the wrist cameras, the top-view RealSense or head-mounted USB camera, the gripper encoder firmware, the foot pedal, and the Meta Quest~3S bridge.
Per-host calibration and provisioning utilities are also included.
A ROS\,2 recording backend captures time-synchronized rosbag2 streams and uploads accepted episodes to S3-compatible storage, and a data processing pipeline written in Rust prepares the dataset by converting the recordings into the LeRobot format with the data-filtering cascade of \Cref{sec:data_proc}.
The processed LeRobot dataset will be released on Hugging Face under the CC-BY-NC-SA-4.0 license.

\subsection{Dataset Release Notes}
YUBI-related datasets, including competition datasets, are released through AIRoA's Hugging Face dataset page: \url{https://huggingface.co/datasets/airoa-org}.

\customparagraph{Format and access}
Releases use the LeRobot v3.0 packed layout: Parquet shards store frame-level data, H.264 MP4 shards store synchronized RGB video, and metadata records the schema, statistics, task vocabulary, episode boundaries, and video offsets. The \texttt{LeRobotDataset} loader resolves these offsets and supports selecting episodes to download only the required shards.

\customparagraph{Episode organization and schema}
Data are synchronized at 30\,Hz. Each stored episode represents a primitive action with a language instruction; a shared execution identifier groups primitive actions into a parent short-horizon task. Episode and parent-task success flags support outcome-based filtering. Features include left/right wrist and center RGB views, gripper joint angles, controller, hand-root, camera, and fingertip poses, tracking-validity and freshness flags, and foot-pedal state. Stored poses use translation and quaternion components $[x,y,z,q_x,q_y,q_z,q_w]$, with absolute and per-step relative transforms in a table-calibrated frame; inter-hand transforms are expressed relative to the left hand. Action fields contain relative controller/hand-root transforms and next-step gripper targets, with the terminal gripper target forward-filled.

\end{document}